\PassOptionsToPackage{table}{xcolor}
\documentclass[11pt]{article}

\usepackage[preprint]{acl}

\usepackage{times}
\usepackage{latexsym}
\usepackage[T1]{fontenc}
\usepackage[utf8]{inputenc}
\usepackage{microtype}
\usepackage{inconsolata}
\usepackage{graphicx}
\usepackage{amsmath}
\usepackage{booktabs}
\usepackage{multirow}
\usepackage{amssymb}

\usepackage{xcolor}
\usepackage[most]{tcolorbox}
\usepackage{tabularx}
\usepackage{array}
\usepackage{enumitem}
\usepackage{fvextra}
\usepackage{caption}
\usepackage{makecell}
\usepackage[normalem]{ulem}
\newcommand{\sweverifier}{\textsc{SWE-Verifier}}

\input{appendix/appendix_style_grouped}

\title{\sweverifier{}: Selective Verification for Cost-Efficient Coding Agents}

\author{\textbf{Kaijian Zou}\textsuperscript{1}\thanks{Corresponding author: \texttt{zkjzou@umich.edu}} \quad
  \textbf{Vaibhav Chennareddy}\textsuperscript{1} \quad
  \textbf{Yunxiang Zhang}\textsuperscript{1} \quad
  \textbf{Naihao Deng}\textsuperscript{1} \\
  \textbf{Longju Bai}\textsuperscript{1} \quad
  \textbf{Jiaxin Pei}\textsuperscript{2} \quad
  \textbf{Rada Mihalcea}\textsuperscript{1} \quad
  \textbf{Lu Wang}\textsuperscript{1} \\[0.5em]
  {\normalfont\textsuperscript{1}University of Michigan \quad \textsuperscript{2}The University of Texas at Austin}}

\begin{document}
\maketitle
\setlength{\parfillskip}{0pt plus .55\columnwidth}

\begin{abstract}
Coding agents spend substantial computation generating actions and processing long interaction histories. Poor action choices can steer agents into low-value states, wasting computation and increasing API costs. Sampling and verifying alternative actions can improve the agent's decisions, but applying this procedure at every step adds substantial overhead. We introduce \sweverifier{}, a lightweight generative model that selectively triggers candidate sampling and verification. It scores the coding agent's proposed action to determine when additional candidate sampling is needed, then ranks the resulting actions before execution. We train \sweverifier{} from Qwen3.5-4B using preference data that groups contrasting actions with their resulting environment observations and rubric-guided teacher scores. Using the mini-swe-agent harness, \sweverifier{} improves resolution from 75.47\% to 77.20\% on MiniMax-M2.7 at 1.09$\times$ cost and generalizes to DeepSeek-V4-Flash without training on its actions, improving resolution from 72.60\% to 74.27\% at 1.13$\times$ cost. Across both models, \sweverifier{} verifies fewer than 4\% of steps and achieves comparable or higher resolution than the trajectory and all-step baselines at substantially lower cost. These results demonstrate that selective candidate sampling and lightweight verification provide complementary gains in coding-agent performance and inference efficiency. Our model, code, and data are available at \url{https://github.com/zkjzou/SWE-Verifier}.
\end{abstract}

\section{Introduction}
\label{sec:introduction}

\begin{figure}[t]
  \centering
  \captionsetup{skip=8pt}
  \includegraphics[viewport=232 152 560 461,clip,width=\columnwidth]{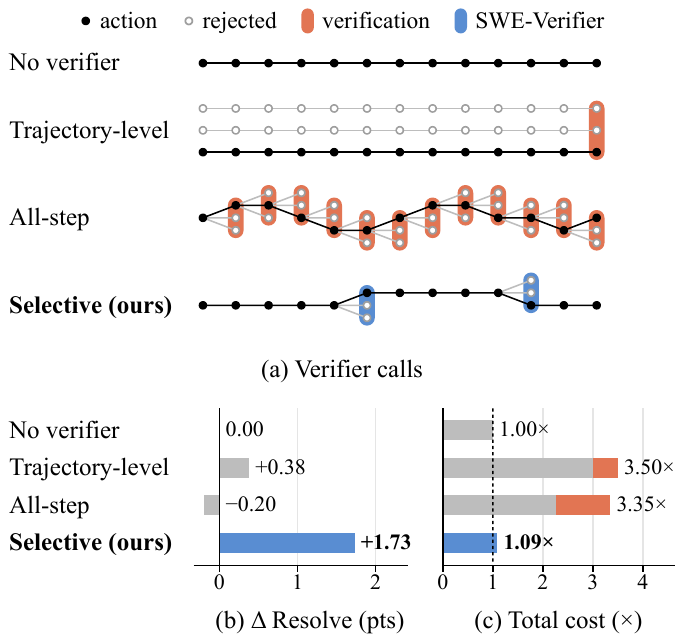}
  \caption{Verification strategies for coding agents on the full 500-task SWE-Bench Verified set. (a) Trajectory-level verification selects among completed rollouts, all-step verification compares candidates at every step, and selective verification compares candidates only at selected steps. (b) Resolution gains and (c) total costs on MiniMax-M2.7: \sweverifier{} improves resolution by 1.73 percentage points at 1.09$\times$ total cost. Both are relative to the no-verifier baseline.}
  \label{fig:verification-comparison}
  \par\vspace{4pt}
\end{figure}

Coding agents powered by large language models (LLMs) solve software-engineering tasks through sequences of actions such as repository inspection, code edits, tool execution, and validation \citep{jimenez2023swebench,yang2024sweagent,wang2024openhands,xia2024agentless}. While the final patch is the ultimate measure of success, the actions leading to it shape both solution quality and computational efficiency. A poor action can steer the agent into a low-value trajectory, requiring additional exploration and corrective work. These detours are expensive: later actions repeatedly process an expanding interaction history, and input and cached-context tokens can account for a substantial portion of coding-agent costs \citep{bai2026aiagentsspendmoney,salim2026tokenomics,wang2025agenttaxo,fan2025swe}. Improving action quality before execution is therefore an opportunity to improve both task success and inference efficiency.

Verification offers a way to allocate additional computation to better decisions. Existing test-time scaling methods apply this idea by adjusting reasoning effort or selecting among independently sampled solutions \citep{snell2025scaling,brown2024largelanguagemonkeys,lightman2023letsverify}. For coding agents, outcome-level and trajectory-level verifiers similarly rank completed patches or rollouts \citep{pan2024swegym,jain2025r2egym,han2026swetrace,zainullina2025guidedsearch,gandhi2025sweprm,dihan2026sweshepherd,shum2025swerm}. However, selection after completion cannot prevent an individual rollout from taking an unproductive branch, and every candidate trajectory must be generated before it can be judged. This motivates verification at intermediate decision points, before the agent commits to its next action.

Step-level verification samples alternative actions from the current agent state and uses a verifier to choose which one to execute (Figure~\ref{fig:verification-comparison}).Existing coding-agent process verifiers often rely on noisy step labels propagated from final outcomes and apply verification at every step \citep{han2026swetrace,zainullina2025guidedsearch,gandhi2025sweprm,dihan2026sweshepherd}. Applying this procedure at every step, however, incurs candidate-generation and scoring overhead throughout the trajectory, typically using the coding agent itself as the verifier. Not every decision warrants this expense. We therefore study \emph{selective step-level verification}: assess the proposed action first, then generate and compare alternatives only when additional scrutiny is useful. The central challenge is \textit{balancing task accuracy against compute budget by deciding which coding-agent actions warrant additional verification.}

We formalize step-level action selection and decompose its policy and verifier costs to compare full and selective verification with trajectory-level scaling.
A preliminary study with MiniMax-M2.7 \citep{minimax2026m27} on SWE-Bench-Verified \citep{jimenez2023swebench} shows that selective verification outperforms all-step verification, achieving similar task performance (76.6\% versus 76.0\%) while significantly reduce the compute cost (1.29$\times$ versus 2.32$\times$ baseline cost) by verifying only 4\% of steps. These observations motivate two complementary design choices to save computing budget: concentrate candidate comparison at useful decision points and offload action evaluation to a lightweight model.

Motivated by these findings, we introduce \emph{\sweverifier{}}, a framework for selective verification of coding-agent actions. A single generative verifier scores the policy's proposed action to determine whether to sample alternatives, then scores the resulting candidates to select the action to execute. We instantiate \sweverifier{} with Qwen3.5-4B \citep{qwen2026qwen35} and train it using an execution-aware preference-data construction recipe. We sample contrasting actions from models with different capabilities, including MiniMax-M2.7 and Qwen3.5-27B, at intermediate states drawn from successful trajectories generated by Opus 4.7 and GPT-5.5. We replay these actions to obtain environment feedback and use GPT-5.6 to produce rubric-guided preference supervision for each candidate group during training.

Our experiments show that selective verification improves the performance--cost tradeoff across two policy models. On MiniMax-M2.7, \sweverifier{} raises resolution from 75.47\% to 77.20\% while verifying only 3.43\% of steps and adding 9\% total cost, outperforming trajectory verification and both evaluated all-step verifiers. Importantly, \sweverifier{} generalizes to DeepSeek-V4-Flash, improving resolution from 72.60\% to 74.27\% at 1.13$\times$ total cost. Together, these results show that selective candidate sampling and lightweight action verification can improve coding-agent performance without the overhead of uniform verification at every step.

\section{Preliminary}
\label{sec:preliminary}

\begin{figure*}[t]
  \centering
  \captionsetup{skip=8pt}
  \includegraphics[viewport=105 8 507 784,clip,angle=-90,width=\textwidth]{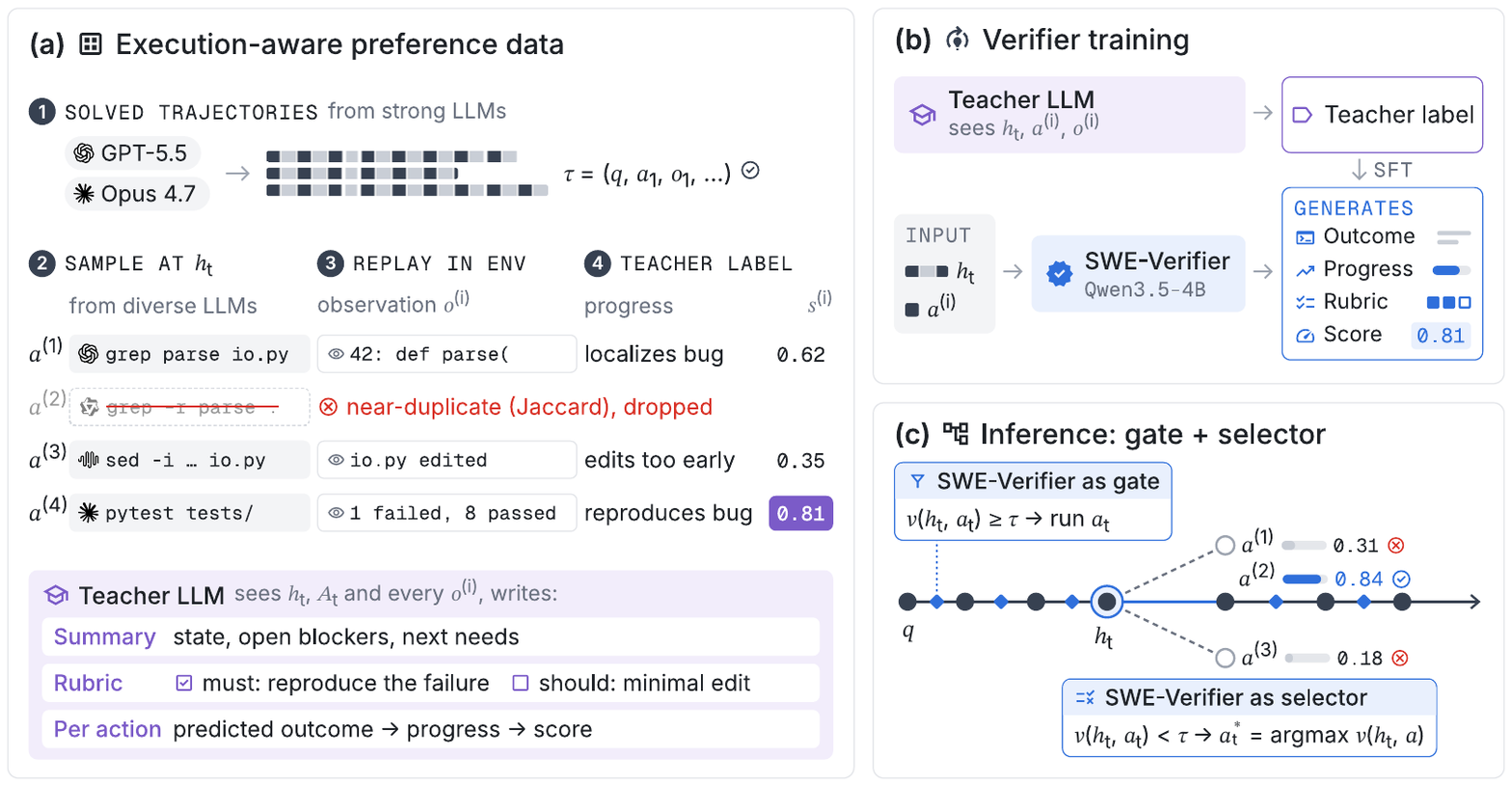}
  \caption{Overview of \sweverifier{}. \textbf{(a)} We construct execution-aware preference data by collecting successful trajectories, sampling and filtering candidate actions, replaying them for environment feedback, and obtaining rubric-guided teacher labels. \textbf{(b)} We train a Qwen3.5-4B verifier to generate outcome predictions, progress assessments, rubric evaluations, and action scores. \textbf{(c)} At inference, the verifier scores the initial action to decide whether to generate alternatives, then ranks candidates when verification is invoked.}
  \label{fig:method-overview}
  \par\vspace{4pt}
\end{figure*}

We formulate step-level verification for coding agents from complementary perspectives on action selection and cost.

\subsection{Conceptual Formulation}
\label{sec:verifier-formulation}
A coding-agent trajectory is $\tau=(q,a_1,o_1,\ldots,a_T,o_T)$, where $q$ is the task context and $a_t$ and $o_t$ are the action and resulting observation at step $t$.

\paragraph{Trajectory-level Verifier.}
Sample $n$ complete trajectories and select one using an outcome verifier $V_{\mathrm{out}}$:
\begin{equation}
    \tau^\star=\arg\max_{\tau_i\in\{\tau_1,\ldots,\tau_n\}}V_{\mathrm{out}}(\tau_i),
    \qquad \tau_i\sim\pi.
\end{equation}
Selection occurs after all $n$ trajectories have been completed, so the verifier cannot guide intermediate actions~\citep{brown2024largelanguagemonkeys,kwok2026llmverifier}.

\paragraph{Step-level Verifier.}
At step $t$, the policy $\pi$ observes history $h_t=(q,a_1,o_1,\ldots,a_{t-1},o_{t-1})$ and generates $n$ candidate actions $A_t=\{a_t^{(1)},\ldots,a_t^{(n)}\}$, with $a_t^{(i)}\sim\pi(\cdot\mid h_t)$. Use an action verifier $v$ to select the next action:
\begin{equation}
    a_t^\star=\arg\max_{a\in A_t}v(h_t,a).
\end{equation}
By comparing alternatives before each action, the verifier can improve local decisions but incurs candidate-generation and scoring overhead throughout the trajectory~\citep{han2026swetrace,zainullina2025guidedsearch,gandhi2025sweprm}.

\paragraph{Selective step-level verifier.}
Because not every step benefits equally from this additional computation, selective verification uses a gate $g_t\in\{0,1\}$ to determine whether to compare candidate actions or execute a policy action directly:
\begin{equation}
    a_t^\star=\begin{cases}
        \arg\max_{a\in A_t}v(h_t,a), & g_t=1,\\
        a_t,\quad a_t\sim\pi(\cdot\mid h_t), & g_t=0.
    \end{cases}
\end{equation}
Selective verification concentrates computation on states where comparing alternatives is expected to be useful. The realized verification rate is $c=\frac{1}{T}\sum_{t=1}^{T}g_t$. The full action-selection derivation is provided in Appendix~\ref{app:action-selection-derivation}.

\subsection{Cost Formulation}
\label{sec:cost-formulation}
Let $x$ and $y$ denote the input and output costs of a baseline trajectory, respectively, such that $C_0=x+y$. We compare the costs of the verification strategies defined above.

\paragraph{Trajectory-level verification.}
Selection occurs after rollout, so the method pays for all $n$ complete trajectories as well as the outcome-verifier cost $C_{\mathrm{out}}$:
\begin{equation}
    C_{\mathrm{traj}}\approx nC_0+C_{\mathrm{out}}.
\end{equation}

\paragraph{Full and selective step-level verification.}
Let $z$ denote the fractional trajectory-length reduction, $\alpha$ the cached-input price multiplier, and $\beta$ the verifier-output cost relative to policy output. Full verification sets $c=1$ and $C_{\mathrm{gate}}=0$; selective verification uses $0\leq c<1$ and incurs $C_{\mathrm{gate}}$. We charge the first candidate the standard input price and later candidates the cached-input price:
\begin{equation}
    \begin{aligned}
        C_{\mathrm{step}}(c)&\approx
        (1-z)\{x[1+\alpha c(n-1)]\\
        &\quad+y[1+c(n-1)]\\
        &\quad+c(x+\beta y)\}+C_{\mathrm{gate}}.
    \end{aligned}
\end{equation}
For the full cost derivation, see Appendix~\ref{app:cost-accounting}; Table~\ref{tab:cost-formulation} summarizes the cost accounting, and Section~\ref{sec:cost-sensitivity} examines pricing sensitivity.

\section{\sweverifier{}: Selective Step-Level Verification for Coding Agents}
\label{sec:swe-verifier}

The formulation in Section~\ref{sec:preliminary} highlights a central challenge: verification is not uniformly beneficial across decision states. At some steps, it provides no gain or can even hurt action selection, while at others its marginal improvement may not justify the added candidate-generation and scoring cost.
Our preliminary study with MiniMax-M2.7 illustrates this tradeoff (Table~\ref{tab:minimax-strategies-full} in Appendix~\ref{sec:minimax-strategies}). Full verification reaches 76.0\% resolution at 2.32$\times$ total cost, with a substantial portion of the overhead coming from using the large policy model to score candidate actions. By contrast, verification restricted to closeout steps reaches 76.6\% resolution while verifying only 4\% of steps at 1.29$\times$ cost.
Periodic verification at a fixed frequency does not provide the same benefit as closeout-only verification, showing that effectiveness depends on which steps are verified. These observations motivate two complementary ways to reduce verification overhead: concentrate verification on steps where it is useful and offload action evaluation to a smaller model. Full results and experimental details appear in Appendix~\ref{sec:minimax-strategies}.

Motivated by these findings, we introduce \emph{\sweverifier{}}, a lightweight generative verifier for selective step-level verification. It scores the policy's proposed action to determine whether additional verification is needed and ranks candidate actions to determine which one the agent should execute. Figure~\ref{fig:method-overview} summarizes the data construction, training, and inference procedure.

To our knowledge, \sweverifier{} is the first framework for coding agents to combine selective step-level verification with a trained generative action verifier. We further introduce an execution-aware construction recipe for its training data: sampling contrasting actions at each trajectory step, replaying them to observe their outcomes, and using a strong teacher to generate rubric-guided preference labels for each candidate group.

\subsection{Gate and Selector Modes}
Selective verification requires two decisions with different roles: whether the expected value of alternatives justifies generating them, and which action to execute once alternatives are available. We assign these decisions to gate and selector modes, respectively.
In \emph{gate mode}, \sweverifier{} scores the policy's initial action $a_t$; the agent executes a high-scoring action directly and invokes additional verification when the score falls below a threshold. In \emph{selector mode}, it scores every action $a\in A_t$ in the candidate group, generating a prediction of each action's likely repository effect and task progress before the agent executes the highest-scoring valid candidate. Thus, the same verifier score determines both when to generate alternatives and which action to select from them. We use Qwen3.5-4B \citep{qwen2026qwen35} as the base model and further improve it through the preference-data construction and training procedure described next.

\subsection{Preference Data Construction}

We construct the training data in four steps.

\paragraph{Step 1: Collect Successful Trajectories.}
We use Opus 4.7 and GPT-5.5 as representative strong coding agents available at the time of our experiments. Together, they provide 500 successful trajectories from the SWE-Smith \texttt{llm\_rewrite} and \texttt{pr\_mirror} instance types.

\paragraph{Step 2: Sample and Filter Candidate Actions.}
For each step in the collected trajectories, we sample five actions from weak and strong models to expose the verifier to contrasting alternatives. We then filter candidates based on their pairwise Jaccard similarity,
\begin{equation}
    J(a,b)=\frac{|S(a)\cap S(b)|}{|S(a)\cup S(b)|}.
\end{equation}
Here, $S(a)$ denotes the set of tokens in action $a$.
We discard candidates whose Jaccard similarity to an already retained action is at least 0.8.
This prevents the teacher from spending labels on near-identical actions.

\paragraph{Step 3: Replay Actions for Environment Feedback.}
We replay each retained candidate from its agent state and record the resulting environment observation. We remove candidates with duplicate non-empty environment observations.

\paragraph{Step 4: Teacher Labeling.} We label candidate steps with GPT-5.6. The teacher receives the current trajectory prefix, candidate actions, and their immediate observations. The teacher then generates structured supervision comprising each action's expected execution outcome, anticipated contribution to task progress, a shared rubric for evaluating the candidate group, and a final score for each candidate. This supervision teaches the verifier to assess not only which action is preferable but also why it is expected to improve the agent state. The teacher prompt is provided in Appendix~\ref{app:prompt-teacher-distillation}, and Appendix~\ref{app:teacher-verifier-case-studies} gives a concrete example of teacher output and analyzes alternative teacher prompts.

\subsection{Training}
\label{sec:verifier-training}

We fine-tune Qwen3.5-4B as the student verifier with a 65,536-token context window on 30k action labels from 6,000 agent steps. The model learns to generate expected execution outcomes, progress assessments, rubric-based evaluations, and final scores for the candidate actions. We train three variants to understand the effects of different training and rubric-generation strategies. \emph{Score Only} uses execution-aware preference training data without rubric generation; \emph{Joint Rubric} generates the rubric and score in one call; and \emph{Two-Stage Rubric} first generates an action-conditioned rubric, then scores the action in a separate call. The two-stage inference prompts are provided in Appendix~\ref{app:prompt-two-stage}.

\section{Experimental Setup}
\label{sec:experimental-setup}

\setcounter{dbltopnumber}{1}
\begin{table*}[t]
\centering
\scriptsize
\setlength{\tabcolsep}{3pt}
\captionsetup{skip=8pt}
\begin{tabular*}{\textwidth}{@{\extracolsep{\fill}}p{0.14\textwidth}p{0.12\textwidth}p{0.22\textwidth}ccc}
\toprule
Policy & Strategy & Verifier Setting & Resolve (\%) $\uparrow$ & Total Cost ($\times$) $\downarrow$ & Steps (Verified \%) \\
\midrule
\multirow{9}{=}{\textbf{MiniMax-M2.7}} & None & No verification & 75.47\textsubscript{\tiny$\pm$1.40} & 1.00 & 46.38 {\tiny (0.00\%)} \\
\addlinespace[2pt]
\cmidrule(l){2-6}
 & \multirow{2}{=}{Trajectory} & LLM-as-a-Verifier@3 & 75.85\textsubscript{\tiny$\pm$0.87} & 3.50 & -- \\
 &  & Oracle@3 & 82.20 & 3.00 & -- \\
\addlinespace[2pt]
\cmidrule(l){2-6}
 & \multirow{2}{=}{All-step} & Self-Verification & 75.27\textsubscript{\tiny$\pm$1.86} & 3.35 & \textbf{38.51} {\tiny (100.00\%)} \\
 &  & Qwen3.5-4B & 76.40\textsubscript{\tiny$\pm$0.69} & 2.26 & \underline{39.98} {\tiny (100.00\%)} \\
\addlinespace[2pt]
\cmidrule(l){2-6}
 & \multirow{4}{=}{Selective} & Untrained & \underline{76.93}\textsubscript{\tiny$\pm$1.21} & 1.20 & 44.05 {\tiny (12.61\%)} \\
 &  & Score Only & \underline{76.93}\textsubscript{\tiny$\pm$0.95} & \textbf{1.05} & 46.93 {\tiny (0.45\%)} \\
 &  & Joint Rubric & 76.87\textsubscript{\tiny$\pm$0.81} & 1.14 & 47.28 {\tiny (3.40\%)} \\
 &  & Two-Stage Rubric & \textbf{77.20}\textsubscript{\tiny$\pm$0.87} & \underline{1.09} & 46.36 {\tiny (3.43\%)} \\
\midrule
\multirow{9}{=}{\textbf{DeepSeek-V4-Flash} (Preview)} & None & No verification & 72.60\textsubscript{\tiny$\pm$0.53} & 1.00 & 41.64 {\tiny (0.00\%)} \\
\addlinespace[2pt]
\cmidrule(l){2-6}
 & \multirow{2}{=}{Trajectory} & LLM-as-a-Verifier@3 & \textbf{74.30}\textsubscript{\tiny$\pm$0.58} & 4.19 & -- \\
 &  & Oracle@3 & 80.60 & 3.00 & -- \\
\addlinespace[2pt]
\cmidrule(l){2-6}
 & \multirow{2}{=}{All-step} & Self-Verification & 72.67\textsubscript{\tiny$\pm$0.90} & 3.99 & \textbf{33.56} {\tiny (100.00\%)} \\
 &  & Qwen3.5-4B & 72.53\textsubscript{\tiny$\pm$0.70} & 2.03 & \underline{37.14} {\tiny (100.00\%)} \\
\addlinespace[2pt]
\cmidrule(l){2-6}
 & \multirow{4}{=}{Selective} & Untrained & 73.60\textsubscript{\tiny$\pm$0.35} & 1.17 & 41.36 {\tiny (8.01\%)} \\
 &  & Score Only & 73.20\textsubscript{\tiny$\pm$0.35} & \underline{1.11} & 41.56 {\tiny (3.94\%)} \\
 &  & Joint Rubric & 73.93\textsubscript{\tiny$\pm$1.03} & \textbf{1.08} & 41.05 {\tiny (3.54\%)} \\
 &  & Two-Stage Rubric & \underline{74.27}\textsubscript{\tiny$\pm$1.01} & 1.13 & 42.15 {\tiny (3.59\%)} \\
\bottomrule
\end{tabular*}
\caption{Verification strategies on the full 500-task SWE-Bench Verified set. DeepSeek-V4-Flash uses correct-high reasoning. Untrained uses the base Qwen3.5-4B model. Steps are mean executed actions, with verification rates in parentheses. Costs are normalized to each policy's no-verifier baseline; e.g., 1.09$\times$ means 9\% more cost than no verification. Among non-baseline, non-oracle settings, bold/underline mark the best/second-best resolution, steps, and costs within each policy.
Across both policies, selective verification improves resolution with substantially lower cost than trajectory-level or all-step verification.
}
\label{tab:student-verifier-results}
\par\vspace{4pt}
\end{table*}

\subsection{Task and Metrics}
\label{sec:task-metrics}

We evaluate the main verification strategies on the full 500-task SWE-Bench Verified set with the mini-swe-agent framework \citep{yang2024sweagent}, using MiniMax-M2.7 \citep{minimax2026m27} and DeepSeek-V4-Flash \citep{deepseekai2026deepseekv4} as policy models. The gate/selector and threshold analyses use a 202-task subset. We report task-resolution rate, average trajectory steps, verify rate, and cost normalized to the corresponding no-verifier policy baseline. For cost accounting, we assume zero inference cost for the local lightweight verifier.

\subsection{Verification Settings}
\label{sec:verifier-settings}

\paragraph{Baselines.}
\emph{No verification} executes the initially sampled policy action and defines the 1.00$\times$ cost reference. \emph{Trajectory verification} samples three complete rollouts: For LLM-as-a-Verifier@3 \citep{kwok2026llmverifier}, the policy model also serves as the verifier that selects among three sampled rollouts; Oracle@3 gives an upper bound by selecting a successful rollout whenever one exists. \emph{All-step verification} generates three candidate actions at every step and selects with either the policy model (Self-Verification) or Qwen3.5-4B \citep{qwen2026qwen35}.

\paragraph{Selective verification.}
\sweverifier{} scores the policy's initial action and triggers additional candidate sampling when its score falls below a threshold, then ranks the candidates and selects the highest-scoring valid action.
We evaluate four variants: \emph{Untrained}, \emph{Score Only}, \emph{Joint Rubric}, and \emph{Two-Stage Rubric}.
We use a 202-task subset with higher outcome variability to identify a cost-effective threshold for each verifier on MiniMax-M2.7. We then apply a common threshold of 0.2 to all DeepSeek-V4-Flash experiments, without separately tuning it for individual verifier variants.

Verifier prompts and additional implementation details are provided in Appendix~\ref{app:verifier-prompts}.

\begin{table*}[t]
  \centering
  \scriptsize
  \setlength{\tabcolsep}{3pt}
  \captionsetup{skip=8pt}
  \begin{tabular*}{\textwidth}{@{\extracolsep{\fill}}p{0.23\textwidth}p{0.23\textwidth}cccc}
    \toprule
    Gate & Selector & Resolve (\%) $\uparrow$ & Steps (Verified \%) & Policy Cost ($\times$) & Total Cost ($\times$) \\
    \midrule
    No verification & -- & 68.48\textsubscript{\tiny$\pm$3.48} & 57.89 {\tiny (0.00\%)} & 1.00 & 1.00 \\
    \midrule
    \multicolumn{6}{@{}l}{\textbf{Threshold $\tau=0.2$}} \\
    \addlinespace[2pt]
    \multirow{2}{=}{\sweverifier{} (Untrained)} & MiniMax-M2.7 & 71.12\textsubscript{\tiny$\pm$3.48} & \textbf{55.50} {\tiny (5.41\%)} & \textbf{1.05} & \underline{1.18} \\
     & \sweverifier{} (Untrained) & 70.63\textsubscript{\tiny$\pm$0.57} & 59.43 {\tiny (5.54\%)} & 1.16 & \textbf{1.16} \\
    \addlinespace[3pt]
    \multirow{2}{=}{\sweverifier{} (Score Only)} & MiniMax-M2.7 & \textbf{73.27}\textsubscript{\tiny$\pm$2.62} & \underline{59.21} {\tiny (5.13\%)} & \underline{1.14} & 1.29 \\
     & \sweverifier{} (Score Only) & \underline{71.95}\textsubscript{\tiny$\pm$3.78} & 61.45 {\tiny (5.79\%)} & 1.24 & 1.24 \\
    \midrule
    \multicolumn{6}{@{}l}{\textbf{Threshold $\tau=0.3$}} \\
    \addlinespace[2pt]
    \multirow{2}{=}{\sweverifier{} (Untrained)} & MiniMax-M2.7 & 70.13\textsubscript{\tiny$\pm$1.03} & 55.49 {\tiny (13.71\%)} & 1.20 & 1.45 \\
     & \sweverifier{} (Untrained) & \underline{71.95}\textsubscript{\tiny$\pm$3.02} & \textbf{53.58} {\tiny (13.33\%)} & \underline{1.11} & \textbf{1.11} \\
    \addlinespace[3pt]
    \multirow{2}{=}{\sweverifier{} (Score Only)} & MiniMax-M2.7 & \textbf{72.61}\textsubscript{\tiny$\pm$2.00} & \underline{54.16} {\tiny (10.47\%)} & \textbf{1.08} & \underline{1.29} \\
     & \sweverifier{} (Score Only) & \underline{71.95}\textsubscript{\tiny$\pm$2.00} & 58.79 {\tiny (11.74\%)} & 1.30 & 1.30 \\
    \bottomrule
  \end{tabular*}
  \caption{Gate/selector ablation for MiniMax-M2.7 on 202 tasks, comparing Untrained and Score Only gates paired with either MiniMax-M2.7 or the corresponding lightweight verifier as selector at two thresholds.
  The trained Score Only gate attains the highest resolution at both thresholds when paired with MiniMax-M2.7, indicating that a lightweight learned gate can reserve a stronger selector for the decisions where it is most useful.
  }
  \label{tab:gate-only-ablation}
  \par\vspace{4pt}
\end{table*}

\begin{figure*}[t]
    \centering
    \begin{minipage}[t]{0.49\textwidth}
        \vspace{0pt}
        \includegraphics[width=\linewidth]{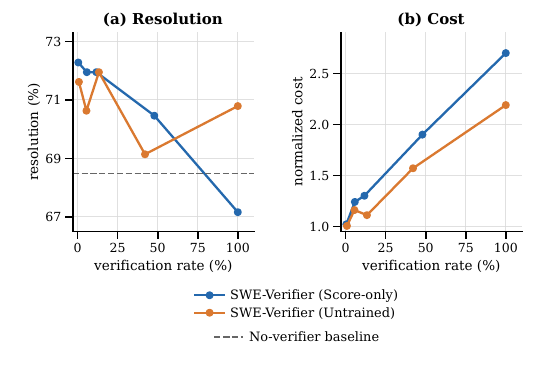}
        \caption{MiniMax-M2.7 threshold sensitivity on 202 tasks at thresholds $\{0.1, 0.2, 0.3, 0.5, 1\}$, where threshold 1 denotes all-step verification: (a) resolution and (b) normalized API cost versus verification rate. Curves show \sweverifier{} (Untrained) and \sweverifier{} (Score-only); the dashed line marks no-verifier resolution. Both variants achieve their highest resolution below the all-step setting, while higher verification rates increase cost without consistently improving resolution.}
        \label{fig:threshold-sensitivity}
    \end{minipage}\hfill
    \begin{minipage}[t]{0.49\textwidth}
        \vspace{0pt}
        \includegraphics[width=\linewidth]{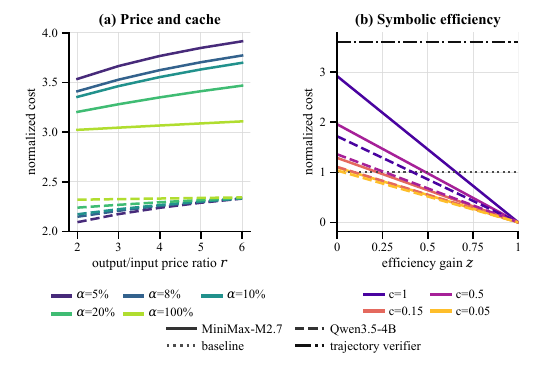}
        \caption{MiniMax-M2.7 cost sensitivity and efficiency plots. (a) Measured cost for all-step verification versus output/input price ratio $r$ and cache multiplier $\alpha$. (b) Formula cost versus fractional trajectory-length reduction $z$ with aggregate input/output cost ratio $\rho=4$, cache multiplier $\alpha=0.2$, and critical-step rate $c\in\{1,0.50,0.15,0.05\}$. All-step verification remains well above baseline cost; selective verification approaches it only with sparse calls and shorter trajectories.}
        \label{fig:cost-sensitivity}
    \end{minipage}
    \par\vspace{4pt}
\end{figure*}

\section{Main Results}
\label{sec:results}

Table~\ref{tab:student-verifier-results} compares verification strategies across two policy models. All \sweverifier{} variants perform both verification triggering and candidate ranking. We focus on resolution and total cost, then examine how training and verification frequency affect this tradeoff.

\paragraph{Selective verification improves resolution with little overhead over the no-verifier baseline.}
On MiniMax-M2.7, the Two-Stage Rubric \sweverifier{} defined in Section~\ref{sec:verifier-training} increases resolution from 75.47\% to 77.20\%, a gain of 1.73 percentage points, at just 1.09$\times$ total cost. It outperforms trajectory-level LLM-as-a-Verifier@3, all-step Self-Verification, and all-step Qwen3.5-4B while verifying only 3.43\% of steps. Selective verification achieves higher resolution at lower cost than the other verification baselines we evaluate, but whether its improvement over no verification is worthwhile depends on the deployment budget.

\paragraph{Training improves the accuracy--cost tradeoff.}
Even the Untrained variant improves MiniMax-M2.7 resolution to 76.93\%. Score Only matches this resolution at 1.05$\times$ cost while verifying only 0.45\% of steps, making it the lowest-cost \sweverifier{} variant on this policy. Joint Rubric reaches 76.87\%, while Two-Stage Rubric achieves the highest resolution, 77.20\%, at 1.09$\times$ cost. Compared with the Untrained variant, the trained variants use verification much less often while maintaining similar or better resolution, indicating that training helps concentrate verification on consequential decisions. Among the trained variants, the stronger result from Two-Stage Rubric suggests that generating the rubric before scoring can further improve action selection.

\paragraph{\sweverifier{} generalizes across policy-model families.}
All four \sweverifier{} variants improve over the 72.60\% no-verifier baseline on DeepSeek-V4-Flash. Two-Stage Rubric performs best among the selective variants, reaching 74.27\% resolution at 1.13$\times$ cost, while Joint Rubric reaches 73.93\% at 1.08$\times$ cost. Two-Stage Rubric nearly matches trajectory-level LLM-as-a-Verifier@3, which reaches 74.30\% at approximately 4.19$\times$ cost, whereas all-step Qwen3.5-4B provides no improvement at 2.03$\times$ cost. These results show that a verifier trained with MiniMax actions can selectively improve a policy from a different model family without the cost of uniform verification.

\paragraph{All-step verification shortens trajectories but increases total cost.}
Executed steps alone understate inference cost because verification may generate and score multiple candidate actions at each step.
\looseness=-1 All-step Self-Verification yields the fewest executed actions on both policies, reducing MiniMax-M2.7 trajectories from 46.38 to 38.51 steps and DeepSeek-V4-Flash trajectories from 41.64 to 33.56 steps. Yet it slightly lowers MiniMax resolution and improves DeepSeek by only 0.07 percentage points, while total costs rise to 3.35$\times$ and 3.99$\times$, respectively. In contrast, Two-Stage Rubric \sweverifier{} keeps trajectory lengths close to the no-verifier baselines while improving resolution at much lower cost. Its advantage therefore comes primarily from avoiding unnecessary candidate generation and verification, rather than from shortening trajectories alone. This distinction motivates the workflow analysis in Section~\ref{sec:workflow-changes}, where we examine whether verification preserves or disrupts the policy's problem-solving process.

\section{Additional Analyses}
\label{sec:verifier-analysis}

\subsection{Gate/Selector Ablation}
\label{sec:gate-selector-ablation}
Table~\ref{tab:gate-only-ablation} compares trained and untrained gates with MiniMax-M2.7 as a fixed selector; Appendix~\ref{app:gate-only-ablation} provides setup details. At $\tau=0.2$, the trained Score Only gate reaches 73.27\% resolution, compared with 71.12\% for the Untrained gate. The trained gate also verifies slightly fewer steps, 5.13\% versus 5.41\%, suggesting that training helps the gate select more useful decision points for verification.

Combining the trained \sweverifier{} gate with the policy model as selector also improves resolution over using \sweverifier{} for both roles: 73.27\% versus 71.95\% at $\tau=0.2$, and 72.61\% versus 71.95\% at $\tau=0.3$. MiniMax-M2.7 is a stronger but much larger selector, so each invocation adds substantial verifier cost. However, its selections produce shorter trajectories and reduce policy-generation cost, partly offsetting this overhead. At $\tau=0.3$, policy cost falls from 1.30$\times$ with the lightweight selector to 1.08$\times$ with MiniMax, leaving their total costs nearly identical at 1.30$\times$ and 1.29$\times$. This suggests a promising direction: use a lightweight trained model to identify critical steps and invoke a stronger selector only when its expected benefit justifies the additional cost.

\subsection{Threshold Sensitivity}
\label{sec:threshold-sensitivity}
To examine how verification frequency affects resolution and cost, we vary the trigger threshold from selective verification to all-step verification.
Figure~\ref{fig:threshold-sensitivity} shows that verifying more steps does not consistently improve resolution. Score Only performs best at its most selective operating point, reaching 72.28\% resolution at 1.02$\times$ cost while verifying only 0.47\% of steps. As its threshold increases and verification becomes more frequent, resolution declines, eventually reaching 67.16\% under all-step verification. The Untrained verifier is less sensitive to this shift, but its best selective setting still outperforms its all-step result.

The larger gap between selective and all-step verification for Score Only suggests that training primarily improves the gate's ability to identify when additional sampling is useful; it does not make every candidate-ranking decision better. A possible explanation is a mismatch between training and inference candidates. Training groups contain diverse actions sampled from both weak and strong models and exclude near duplicates, whereas inference candidates come from the same policy and may differ only subtly. This mismatch may limit the trained verifier's ability to distinguish candidates when it is applied at every step.

\subsection{Cost Sensitivity}
\label{sec:cost-sensitivity}
\paragraph{Pricing and serving effects.}
Figure~\ref{fig:cost-sensitivity}(a) shows that all-step verification remains more expensive than the no-verifier baseline across pricing settings. Self-verification costs $3.0$--$3.9\times$ the baseline because it incurs both candidate-generation and verifier-input costs; stronger cache discounts increase its normalized cost because the baseline benefits more. All-step verification with Qwen3.5-4B costs $2.1$--$2.3\times$ and benefits more from caching because the local verifier adds no API input cost. Both become more expensive as output prices rise relative to input prices because candidate generation accounts for more of the total cost.

\paragraph{Break-even condition.}
Figure~\ref{fig:cost-sensitivity}(b) shows the trajectory reduction needed to match baseline cost with $\rho=x/y=4$, $\alpha=0.2$, and $\beta=2$. As $c$ rises from 0.05 to 1, the required reduction rises from 9\% to 66\% for self-verification and from 3\% to 42\% without API verifier cost. At $c=0.15$, these values are 22\% and 10\%, compared with the largest observed reductions of 17\% for MiniMax-M2.7 and 19\% for DeepSeek-V4-Flash (Table~\ref{tab:student-verifier-results}). Thus, selective verification approaches baseline cost only with sparse calls or sufficient trajectory shortening.

\subsection{Workflow Changes}
\label{sec:workflow-changes}

All-step verification may provide limited gains when it changes how the policy approaches the task. We examine this failure mode through verifier-as-tool, a setting in which the policy knows that a verifier will select among its proposed actions and is explicitly prompted to generate diverse alternatives. Figure~\ref{fig:phase-position-single-column} shows that ordinary step-level verification largely preserves the baseline phase structure, with phase agreement above 93\% throughout the trajectory. Verifier-as-tool diverges most during the early transition from exploration to fixing, and its resolution falls below the baseline despite producing shorter trajectories. This suggests that verification is harmful when the requirement to generate alternatives interferes with the policy's exploration and repair process. Effective verification should improve selected decisions while preserving the policy's broader workflow. Appendix~\ref{sec:minimax-strategies} reports the full MiniMax-M2.7 comparison, while Appendix~\ref{app:minimax-astropy-14369} analyzes a post-validation final-patch failure.

\section{Related Work}
\label{sec:related-work}

\paragraph{Coding Agents.}
SWE-bench evaluates repository-level issue resolution \citep{jimenez2023swebench}, motivating coding agents built around interactive tool use, structured repair pipelines, and trajectory search \citep{yang2024sweagent,wang2024openhands,zhang2024autocoderover,xia2024agentless,antoniades2024swesearch}. We study verification of these agents' intermediate decisions about what to inspect, execute, edit, and submit.

\paragraph{Test-time Scaling.}
Verifiers support test-time scaling by scoring sampled solutions or intermediate steps, including through process supervision and generative reward modeling \citep{lightman2023letsverify,chen2025rmr1,khalifa2025thinkprm}. For interactive agents, these scores can assess progress toward task completion \citep{xi2025agentprm,chae2025webshepherd,zhang2025progrm,choudhury2025agentprmframework}.
For coding agents, SWE-Gym and R2E-Gym train agents and verifiers in executable environments \citep{pan2024swegym,jain2025r2egym}; SWE-TRACE and Guided Search guide search using rubric-conditioned process rewards or learned action values \citep{han2026swetrace,zainullina2025guidedsearch}. Other work studies trajectory reranking, course correction, action-level process rewards, and execution-free evaluation \citep{wang2026rubric,gandhi2025sweprm,dihan2026sweshepherd,shum2025swerm}. \sweverifier{} combines a trained generative action verifier with selective invocation, using action scores both to identify steps that warrant additional verification and to rank candidate actions before execution.

\paragraph{Inference Efficiency and Selective Verification.}
Repeated sampling increases inference cost \citep{brown2024largelanguagemonkeys}, motivating model cascades, cost-aware routing, and adaptive compute allocation \citep{chen2023frugalgpt,ong2024routellm,manvi2024adaptivecompute,chen2025vgsearch,liu2026confspec,alomrani2025reasoning}. Coding agents similarly use budget-aware tool use and efficient scaffolds to control long-horizon execution costs \citep{wang2025efficient,gao2025more,liu2025budget,lin2026stop}. Selective verification reduces overhead by avoiding additional candidate generation and scoring at every step, saving policy and verifier computation as well as repeated processing of long trajectory histories. Distilling teacher supervision into a smaller verifier further lowers evaluation cost \citep{hsieh-etal-2023-distilling}. Together, selective invocation and lightweight scoring target complementary sources of verification overhead: the frequency of candidate comparison and the cost of each evaluation.

\section{Conclusion}
\label{sec:conclusion}
We introduced \sweverifier{}, a lightweight verifier that selectively samples and ranks alternative coding-agent actions. Trained with execution-aware teacher supervision, it improves MiniMax-M2.7 resolution from 75.47\% to 77.20\% while verifying 3.43\% of steps at 1.09$\times$ total cost. It generalizes to DeepSeek-V4-Flash, improving resolution from 72.60\% to 74.27\% at 1.13$\times$ cost. Our analyses show that verification works best selectively: a gate targets decisions that benefit from verification, whereas uniform verification adds substantial cost without consistent gains. Selective verification thus improves coding agents while controlling inference overhead.

\newpage
\section*{Limitations}

Our experiments use the mini-swe-agent harness to evaluate MiniMax-M2.7 and DeepSeek-V4-Flash on SWE-Bench-Verified. The observed accuracy--cost tradeoffs may differ for other coding benchmarks, repository distributions, serving systems, or agent frameworks, especially when trajectories have different lengths or when candidate actions are not generated from a shared prompt.

\looseness=-1 Our cost analysis also depends on practical serving assumptions. Step-level verification is most favorable when repeated candidate prompts receive cache discounts and verifier calls are inexpensive relative to policy calls. Providers with different cache discounts or batching behavior may change the break-even point. Finally, \sweverifier{} is trained from teacher-labeled action preferences and successful trajectory states, so its judgments may inherit teacher biases and may be less reliable on failure modes or domains underrepresented in the training data.

\setlength{\parfillskip}{0pt plus 1fil}
 \bibliography{custom}

@inproceedings{jimenez2023swebench,
  title = {{SWE-bench}: Can Language Models Resolve Real-World {GitHub} Issues?},
  author = {Jimenez, Carlos E. and Yang, John and Wettig, Alexander and Yao, Shunyu and Pei, Kexin and Press, Ofir and Narasimhan, Karthik},
  booktitle = {International Conference on Learning Representations},
  year = {2024},
  url = {https://arxiv.org/abs/2310.06770}
}

@inproceedings{yang2024sweagent,
  title = {{SWE-agent}: Agent-Computer Interfaces Enable Automated Software Engineering},
  author = {Yang, John and Jimenez, Carlos E. and Wettig, Alexander and Lieret, Kilian and Yao, Shunyu and Narasimhan, Karthik and Press, Ofir},
  booktitle = {Advances in Neural Information Processing Systems},
  year = {2024},
  url = {https://arxiv.org/abs/2405.15793}
}

@inproceedings{wang2024openhands,
  title = {{OpenHands}: An Open Platform for AI Software Developers as Generalist Agents},
  author = {Wang, Xingyao and Li, Boxuan and Song, Yufan and Xu, Frank F. and Tang, Xiangru and Zhuge, Mingchen and Pan, Jiayi and Song, Yueqi and Li, Bowen and Singh, Jaskirat and Tran, Hoang H. and Li, Fuqiang and Ma, Ren and Zheng, Mingzhang and Qian, Bill and Shao, Yanjun and Muennighoff, Niklas and Zhang, Yizhe and Hui, Binyuan and Lin, Junyang and Brennan, Robert and Peng, Hao and Ji, Heng and Neubig, Graham},
  booktitle = {International Conference on Learning Representations},
  year = {2025},
  url = {https://arxiv.org/abs/2407.16741}
}

@inproceedings{zhang2024autocoderover,
  title = {{AutoCodeRover}: Autonomous Program Improvement},
  author = {Zhang, Yuntong and Ruan, Haifeng and Fan, Zhiyu and Roychoudhury, Abhik},
  booktitle = {Proceedings of the 33rd ACM SIGSOFT International Symposium on Software Testing and Analysis},
  year = {2024},
  url = {https://arxiv.org/abs/2404.05427}
}

@inproceedings{xia2024agentless,
  title = {Agentless: Demystifying {LLM}-based Software Engineering Agents},
  author = {Xia, Chunqiu Steven and Deng, Yinlin and Dunn, Soren and Zhang, Lingming},
  booktitle = {Proceedings of the ACM on Software Engineering},
  year = {2025},
  url = {https://arxiv.org/abs/2407.01489}
}

@inproceedings{antoniades2024swesearch,
  title = {{SWE-Search}: Enhancing Software Agents with Monte Carlo Tree Search and Iterative Refinement},
  author = {Antoniades, Antonis and {\"O}rwall, Albert and Zhang, Kexun and Xie, Yuxi and Goyal, Anirudh and Wang, William Yang},
  booktitle = {International Conference on Learning Representations},
  year = {2025},
  url = {https://arxiv.org/abs/2410.20285}
}

@misc{brown2024largelanguagemonkeys,
  title = {Large Language Monkeys: Scaling Inference Compute with Repeated Sampling},
  author = {Brown, Bradley and Juravsky, Jordan and Ehrlich, Ryan and Clark, Ronald and Le, Quoc V. and R{\'e}, Christopher and Mirhoseini, Azalia},
  year = {2024},
  eprint = {2407.21787},
  archivePrefix = {arXiv},
  primaryClass = {cs.LG},
  url = {https://arxiv.org/abs/2407.21787}
}

@article{chen2023frugalgpt,
  title = {{FrugalGPT}: How to Use Large Language Models While Reducing Cost and Improving Performance},
  author = {Chen, Lingjiao and Zaharia, Matei and Zou, James},
  journal = {Transactions on Machine Learning Research},
  year = {2023},
  url = {https://arxiv.org/abs/2305.05176}
}

@misc{ong2024routellm,
  title = {{RouteLLM}: Learning to Route {LLM}s with Preference Data},
  author = {Ong, Isaac and Almahairi, Amjad and Wu, Vincent and Chiang, Wei-Lin and Wu, Tianhao and Gonzalez, Joseph E. and Kadous, M. Waleed and Stoica, Ion},
  year = {2024},
  eprint = {2406.18665},
  archivePrefix = {arXiv},
  primaryClass = {cs.CL},
  url = {https://arxiv.org/abs/2406.18665}
}

@misc{manvi2024adaptivecompute,
  title = {Adaptive Inference-Time Compute: {LLM}s Can Predict if They Can Do Better, Even Mid-Generation},
  author = {Manvi, Rohin and Singh, Anikait and Ermon, Stefano},
  year = {2024},
  eprint = {2410.02725},
  archivePrefix = {arXiv},
  primaryClass = {cs.LG},
  url = {https://arxiv.org/abs/2410.02725}
}

@inproceedings{chen2025vgsearch,
  title = {Rethinking Optimal Verification Granularity for Compute-Efficient Test-Time Scaling},
  author = {Chen, Hao Mark and Lu, Guanxi and Okoshi, Yasuyuki and Mo, Zhiwen and Motomura, Masato and Fan, Hongxiang},
  booktitle = {Advances in Neural Information Processing Systems},
  year = {2025},
  url = {https://arxiv.org/abs/2505.11730}
}

@misc{liu2026confspec,
  title = {{ConfSpec}: Efficient Step-Level Speculative Reasoning via Confidence-Gated Verification},
  author = {Liu, Siran and He, Cyril Y.},
  year = {2026},
  eprint = {2602.18447},
  archivePrefix = {arXiv},
  primaryClass = {cs.CL},
  url = {https://arxiv.org/abs/2602.18447}
}

@inproceedings{lightman2023letsverify,
  title = {Let's Verify Step by Step},
  author = {Lightman, Hunter and Kosaraju, Vineet and Burda, Yura and Edwards, Harri and Baker, Bowen and Lee, Teddy and Leike, Jan and Schulman, John and Sutskever, Ilya and Cobbe, Karl},
  booktitle = {International Conference on Learning Representations},
  year = {2024},
  url = {https://arxiv.org/abs/2305.20050}
}

@misc{chen2025rmr1,
  title = {{RM-R1}: Reward Modeling as Reasoning},
  author = {Chen, Xiusi and Li, Gaotang and Wang, Ziqi and Jin, Bowen and Qian, Cheng and Wang, Yu and Wang, Hongru and Zhang, Yu and Zhang, Denghui and Zhang, Tong and Tong, Hanghang and Ji, Heng},
  year = {2025},
  eprint = {2505.02387},
  archivePrefix = {arXiv},
  primaryClass = {cs.CL},
  url = {https://arxiv.org/abs/2505.02387}
}

@article{khalifa2025thinkprm,
  title = {Process Reward Models That Think},
  author = {Khalifa, Muhammad and Agarwal, Rishabh and Logeswaran, Lajanugen and Kim, Jaekyeom and Peng, Hao and Lee, Moontae and Lee, Honglak and Wang, Lu},
  journal = {Transactions on Machine Learning Research},
  year = {2026},
  url = {https://arxiv.org/abs/2504.16828}
}

@misc{xi2025agentprm,
  title = {{AgentPRM}: Process Reward Models for {LLM} Agents via Step-Wise Promise and Progress},
  author = {Xi, Zhiheng and Liao, Chenyang and Li, Guanyu and Yang, Yajie and Chen, Wenxiang and Zhang, Zhihao and Wang, Binghai and Jin, Senjie and Zhou, Yuhao and Guan, Jian and Wu, Wei and Ji, Tao and Gui, Tao and Zhang, Qi and Huang, Xuanjing},
  year = {2025},
  eprint = {2511.08325},
  archivePrefix = {arXiv},
  primaryClass = {cs.CL},
  url = {https://arxiv.org/abs/2511.08325}
}

@misc{chae2025webshepherd,
  title = {Web-Shepherd: Advancing {PRM}s for Reinforcing Web Agents},
  author = {Chae, Hyungjoo and Kim, Sunghwan and Cho, Junhee and Kim, Seungone and Moon, Seungjun and Hwangbo, Gyeom and Lim, Dongha and Kim, Minjin and Hwang, Yeonjun and Gwak, Minju and Choi, Dongwook and Kang, Minseok and Im, Gwanhoon and Cho, ByeongUng and Kim, Hyojun and Han, Jun Hee and Kwon, Taeyoon and Kim, Minju and Kwak, Beong-woo and Kang, Dongjin and Yeo, Jinyoung},
  year = {2025},
  eprint = {2505.15277},
  archivePrefix = {arXiv},
  primaryClass = {cs.CL},
  url = {https://arxiv.org/abs/2505.15277}
}

@misc{zhang2025progrm,
  title = {{ProgRM}: Build Better GUI Agents with Progress Rewards},
  author = {Zhang, Danyang and Zhang, Situo and Yang, Ziyue and Zhu, Zichen and Zhao, Zihan and Cao, Ruisheng and Chen, Lu and Yu, Kai},
  year = {2025},
  eprint = {2505.18121},
  archivePrefix = {arXiv},
  primaryClass = {cs.AI},
  url = {https://arxiv.org/abs/2505.18121}
}

@misc{choudhury2025agentprmframework,
  title = {Process Reward Models for {LLM} Agents: Practical Framework and Directions},
  author = {Choudhury, Sanjiban},
  year = {2025},
  eprint = {2502.10325},
  archivePrefix = {arXiv},
  primaryClass = {cs.AI},
  url = {https://arxiv.org/abs/2502.10325}
}

@inproceedings{pan2024swegym,
  title = {Training Software Engineering Agents and Verifiers with {SWE-Gym}},
  author = {Pan, Jiayi and Wang, Xingyao and Neubig, Graham and Jaitly, Navdeep and Ji, Heng and Suhr, Alane and Zhang, Yizhe},
  booktitle = {International Conference on Machine Learning},
  year = {2025},
  url = {https://arxiv.org/abs/2412.21139}
}

@misc{jain2025r2egym,
  title = {{R2E-Gym}: Procedural Environments and Hybrid Verifiers for Scaling Open-Weights {SWE} Agents},
  author = {Jain, Naman and Singh, Jaskirat and Shetty, Manish and Zheng, Liang and Sen, Koushik and Stoica, Ion},
  year = {2025},
  eprint = {2504.07164},
  archivePrefix = {arXiv},
  primaryClass = {cs.SE},
  url = {https://arxiv.org/abs/2504.07164}
}

@misc{han2026swetrace,
  title = {{SWE-TRACE}: Optimizing Long-Horizon {SWE} Agents Through Rubric Process Reward Models and Heuristic Test-Time Scaling},
  author = {Han, Hao and Xie, Jin and Ma, Xuehao and Zhu, Weiquan and Zhang, Ziyao and Long, ZhiLiang and Chen, Hongkai and Ye, Qingwen},
  year = {2026},
  eprint = {2604.14820},
  archivePrefix = {arXiv},
  primaryClass = {cs.SE},
  url = {https://arxiv.org/abs/2604.14820}
}

@inproceedings{zainullina2025guidedsearch,
  title = {Guided Search Strategies in Non-Serializable Environments with Applications to Software Engineering Agents},
  author = {Zainullina, Karina and Golubev, Alexander and Trofimova, Maria and Polezhaev, Sergei and Badertdinov, Ibragim and Litvintseva, Daria and Karasik, Simon and Fisin, Filipp and Skvortsov, Sergei and Nekrashevich, Maksim and Shevtsov, Anton and Yangel, Boris},
  booktitle = {Proceedings of the 42nd International Conference on Machine Learning},
  year = {2025},
  url = {https://arxiv.org/abs/2505.13652}
}

@misc{gandhi2025sweprm,
  title = {When Agents go Astray: Course-Correcting {SWE} Agents with {PRM}s},
  author = {Gandhi, Shubham and Tsay, Jason and Ganhotra, Jatin and Kate, Kiran and Rizk, Yara},
  year = {2025},
  eprint = {2509.02360},
  archivePrefix = {arXiv},
  primaryClass = {cs.SE},
  url = {https://arxiv.org/abs/2509.02360}
}

@misc{dihan2026sweshepherd,
  title = {{SWE-Shepherd}: Advancing {PRM}s for Reinforcing Code Agents},
  author = {Dihan, Mahir Labib and Khan, Md Ashrafur Rahman},
  year = {2026},
  eprint = {2604.10493},
  archivePrefix = {arXiv},
  primaryClass = {cs.SE},
  url = {https://arxiv.org/abs/2604.10493}
}

@misc{shum2025swerm,
  title = {{SWE-RM}: Execution-free Feedback for Software Engineering Agents},
  author = {Shum, KaShun and Hui, Binyuan and Chen, Jiawei and Zhang, Lei and {X. W.} and Yang, Jiaxi and Huang, Yuzhen and Lin, Junyang and He, Junxian},
  year = {2025},
  eprint = {2512.21919},
  archivePrefix = {arXiv},
  primaryClass = {cs.SE},
  url = {https://arxiv.org/abs/2512.21919}
}

@misc{bai2026aiagentsspendmoney,
      title={How Do AI Agents Spend Your Money? Analyzing and Predicting Token Consumption in Agentic Coding Tasks}, 
      author={Longju Bai and Zhemin Huang and Xingyao Wang and Jiao Sun and Rada Mihalcea and Erik Brynjolfsson and Alex Pentland and Jiaxin Pei},
      year={2026},
      eprint={2604.22750},
      archivePrefix={arXiv},
      primaryClass={cs.CL},
      url={https://arxiv.org/abs/2604.22750}, 
}

@inproceedings{hsieh-etal-2023-distilling,
    title = "Distilling Step-by-Step! Outperforming Larger Language Models with Less Training Data and Smaller Model Sizes",
    author = "Hsieh, Cheng-Yu  and
      Li, Chun-Liang  and
      Yeh, Chih-kuan  and
      Nakhost, Hootan  and
      Fujii, Yasuhisa  and
      Ratner, Alex  and
      Krishna, Ranjay  and
      Lee, Chen-Yu  and
      Pfister, Tomas",
    editor = "Rogers, Anna  and
      Boyd-Graber, Jordan  and
      Okazaki, Naoaki",
    booktitle = "Findings of the Association for Computational Linguistics: ACL 2023",
    month = jul,
    year = "2023",
    address = "Toronto, Canada",
    publisher = "Association for Computational Linguistics",
    url = "https://aclanthology.org/2023.findings-acl.507/",
    doi = "10.18653/v1/2023.findings-acl.507",
    pages = "8003--8017"
}

@article{salim2026tokenomics,
  title={Tokenomics: Quantifying Where Tokens Are Used in Agentic Software Engineering},
  author={Salim, Mohamad and Latendresse, Jasmine and Khatoonabadi, SayedHassan and Shihab, Emad},
  journal={arXiv preprint arXiv:2601.14470},
  year={2026}
}

@inproceedings{
wang2025agenttaxo,
title={AgentTaxo: Dissecting and Benchmarking Token Distribution of {LLM} Multi-Agent Systems},
author={Qian Wang and Zhenheng Tang and ZICHEN JIANG and Nuo Chen and Tianyu Wang and Bingsheng He},
booktitle={ICLR 2025 Workshop on Foundation Models in the Wild},
year={2025},
url={https://openreview.net/forum?id=0iLbiYYIpC}
}

@article{fan2025swe,
  title={Swe-effi: Re-evaluating software ai agent system effectiveness under resource constraints},
  author={Fan, Zhiyu and Vasilevski, Kirill and Lin, Dayi and Chen, Boyuan and Chen, Yihao and Zhong, Zhiqing and Zhang, Jie M and He, Pinjia and Hassan, Ahmed E},
  journal={arXiv preprint arXiv:2509.09853},
  year={2025}
}

@article{gao2025more,
  title={More with less: An empirical study of turn-control strategies for efficient coding agents},
  author={Gao, Pengfei and Peng, Chao},
  journal={arXiv preprint arXiv:2510.16786},
  year={2025}
}

@article{liu2025budget,
  title={Budget-aware tool-use enables effective agent scaling},
  author={Liu, Tengxiao and Wang, Zifeng and Miao, Jin and Hsu, I and Yan, Jun and Chen, Jiefeng and Han, Rujun and Xu, Fangyuan and Chen, Yanfei and Jiang, Ke and others},
  journal={arXiv preprint arXiv:2511.17006},
  year={2025}
}

@inproceedings{
lin2026stop,
title={Stop Wasting Your Tokens: Towards Efficient Runtime Multi-Agent Systems},
author={Fulin Lin and Shaowen Chen and Ruishan Fang and Hongwei Wang and Tao Lin},
booktitle={The Fourteenth International Conference on Learning Representations},
year={2026},
url={https://openreview.net/forum?id=pzFhtpkabh}
}

@inproceedings{
    snell2025scaling,
    title={Scaling {LLM} Test-Time Compute Optimally Can be More Effective than Scaling Parameters for Reasoning},
    author={Charlie Victor Snell and Jaehoon Lee and Kelvin Xu and Aviral Kumar},
    booktitle={The Thirteenth International Conference on Learning Representations},
    year={2025},
    url={https://openreview.net/forum?id=4FWAwZtd2n}
}

@article{wang2025efficient,
  title={Efficient agents: Building effective agents while reducing cost},
  author={Wang, Ningning and Hu, Xavier and Liu, Pai and Zhu, He and Hou, Yue and Huang, Heyuan and Zhang, Shengyu and Yang, Jian and Liu, Jiaheng and Zhang, Ge and others},
  journal={arXiv preprint arXiv:2508.02694},
  year={2025}
}

@article{alomrani2025reasoning,
  title={Reasoning on a budget: A survey of adaptive and controllable test-time compute in llms},
  author={Alomrani, Mohammad Ali and Zhang, Yingxue and Li, Derek and Sun, Qianyi and Pal, Soumyasundar and Zhang, Zhanguang and Hu, Yaochen and Ajwani, Rohan Deepak and Valkanas, Antonios and Karimi, Raika and others},
  journal={arXiv preprint arXiv:2507.02076},
  year={2025}
}

@article{wang2026rubric,
  title={A Rubric-Supervised Critic from Sparse Real-World Outcomes},
  author={Wang, Xingyao and Chen, Valerie and Ji, Heng and Neubig, Graham},
  journal={arXiv preprint arXiv:2603.03800},
  year={2026}
}

@misc{minimax2026m27,
  title        = {{MiniMax M2.7}: Early Echoes of Self-Evolution},
  author       = {{MiniMax}},
  year         = {2026},
  month        = mar,
  url          = {https://www.minimax.io/news/minimax-m27-en},
  note         = {Official release page}
}

@misc{deepseekai2026deepseekv4,
  title        = {{DeepSeek-V4}: Towards Highly Efficient Million-Token Context Intelligence},
  author       = {{DeepSeek-AI}},
  year         = {2026},
  url          = {https://huggingface.co/deepseek-ai/DeepSeek-V4-Flash},
  note         = {Official model card}
}

@misc{qwen2026qwen35,
  title        = {{Qwen3.5}: Towards Native Multimodal Agents},
  author       = {{Qwen Team}},
  year         = {2026},
  month        = feb,
  url          = {https://qwen.ai/blog?id=qwen3.5},
  note         = {Official release page}
}

@misc{kwok2026llmverifier,
  title  = {{LLM-as-a-Verifier}: A General-Purpose Verification Framework},
  author = {Kwok, Jacky and Li, Shulu and Atreya, Pranav and Liu, Yuejiang and Pavone, Marco and Stoica, Ion and Mirhoseini, Azalia},
  year   = {2026},
  note   = {Notion blog}
}

\clearpage
\appendix

\section{MiniMax-M2.7 Verification Strategies}
\label{sec:minimax-strategies}

We evaluate MiniMax-M2.7 on SWE-Bench-Verified across trajectory-level, step-level, selective-step, and small-verifier settings. Table~\ref{tab:minimax-strategies-full} reports the full comparison. We report task resolution, trajectory length, verification rate, and cost normalized to the three-seed no-verifier MiniMax-M2.7 baseline. Verification rate is the fraction of steps where the verifier is invoked; candidate actions are verified only when at least two are unique.

\paragraph{Verifier settings.}
The trajectory-level verifier samples multiple complete rollouts and uses an LLM to select the final trajectory. Step-level verifiers instead select among candidate actions within a rollout: \emph{All Steps} variants verify every state, whereas \emph{Selective Steps} variants verify only a subset based on step frequency or an agent phase classified by Qwen3.5-27B. \emph{Verifier-as-tool} allows the policy to call a verifier tool with multiple candidate actions. The small-model setting keeps the policy fixed while replacing the action verifier with Qwen3.5-4B.

\begin{table*}[!t]
\centering
\scriptsize
\setlength{\tabcolsep}{3pt}
\captionsetup{skip=8pt}
\begin{tabular*}{\textwidth}{@{\extracolsep{\fill}}p{0.38\textwidth}cccccc}
\toprule
{} & Resolve & $\Delta$ Resolve & Steps & Verify Rate & Policy Cost & Total Cost \\
\midrule
MiniMax-M2.7 baseline & 75.5\% & +0.0 & 46.4 & 0\% & 1.00$\times$ & 1.00$\times$ \\
\cmidrule(lr){1-7}
\multicolumn{7}{@{}l}{\textbf{\textit{Trajectory-level upper bound}}} \\
\quad Oracle@3 & 82.2\% & +6.7 & 139.1 & -- & 3.00$\times$ & 3.00$\times$ \\
\cmidrule(lr){1-7}
\multicolumn{7}{@{}l}{\textbf{\textit{Step-level verifier}}} \\
\addlinespace[1pt]
\multicolumn{7}{@{}l}{\quad \textbf{All Steps}} \\
\qquad verifier ($n=3$) & 76.2\% & +0.7 & 38.2 & 100\% & 1.15$\times$ & 2.16$\times$ \\
\qquad verifier ($n=5$) & 76.0\% & +0.5 & 35.9 & 100\% & 1.38$\times$ & 2.32$\times$ \\
\qquad verifier prefix ($n=3$) & 75.6\% & +0.1 & 35.7 & 89\% & 1.00$\times$ & 1.77$\times$ \\
\addlinespace[1pt]
\multicolumn{7}{@{}l}{\quad \textbf{Verifier-as-Tool}} \\
\qquad verifier-as-tool ($n=3$) & 70.2\% & -5.3 & 30.4 & 99\% & 0.71$\times$ & 1.40$\times$ \\
\qquad verifier-as-tool ($n=5$) & 69.2\% & -6.3 & 37.1 & 99\% & 0.99$\times$ & 1.97$\times$ \\
\addlinespace[1pt]
\multicolumn{7}{@{}l}{\quad \textbf{Selective Steps} ($n=5$)} \\
\qquad every 5 steps & 75.0\% & -0.5 & 44.2 & 21\% & 1.08$\times$ & 1.39$\times$ \\
\qquad explore phase & 76.0\% & +0.5 & 39.3 & 41\% & 1.28$\times$ & 1.68$\times$ \\
\qquad validate phase & 76.4\% & +0.9 & 43.3 & 33\% & 1.47$\times$ & 1.97$\times$ \\
\qquad fix phase & 76.2\% & +0.7 & 45.3 & 9\% & 1.16$\times$ & 1.35$\times$ \\
\qquad closeout phase & 76.6\% & +1.1 & 47.7 & 4\% & 1.16$\times$ & 1.29$\times$ \\
\cmidrule(lr){1-7}
\multicolumn{7}{@{}l}{\textbf{\textit{Small-model verifier}}} \\
\quad Qwen3.5-4B ($n=3$) & 76.0\% & +0.5 & 40.8 & 89\% & 1.28$\times$ & 1.28$\times$ \\
\bottomrule
\end{tabular*}
\caption{MiniMax-M2.7 results on the full 500-task SWE-Bench Verified set. Each step-level and small-model verification setting reports a single run on all 500 tasks. The no-verifier baseline averages three runs, and Oracle@3 selects a successful rollout whenever one exists among those three runs. $\Delta$ Resolve is measured against the averaged no-verifier baseline. Verify Rate is the fraction of trajectory steps where candidate actions are verified. Policy Cost counts only MiniMax-M2.7 policy tokens; Total Cost includes verifier-selection calls but excludes label-gate cost for selective-step rows. Small trained verifier calls are treated as near-zero relative to frontier policy calls, so their Total Cost equals Policy Cost.}
\label{tab:minimax-strategies-full}
\end{table*}

\subsection{Full Verification Improves Step Efficiency but Remains Costly}
\label{sec:step-level-efficiency}

The baseline solves 75.5\% of tasks with 46.4 steps on average. Full verification with five candidates slightly raises resolution to 76.0\% and shortens the average trajectory to 35.9 steps, but its total cost reaches 2.32$\times$ the baseline because candidate generation and verifier scoring occur throughout the trajectory. With three candidates, full verification reaches 76.2\% resolution in 38.2 steps at 2.16$\times$ total cost. These results suggest that step-level verification can improve local decisions and reduce environment interactions, but the savings from shorter trajectories do not offset the cost of verifying every step.

\subsection{Verifier Awareness Can Disrupt the Workflow}

\begin{figure}[t]
  \centering
  \includegraphics[width=\columnwidth]{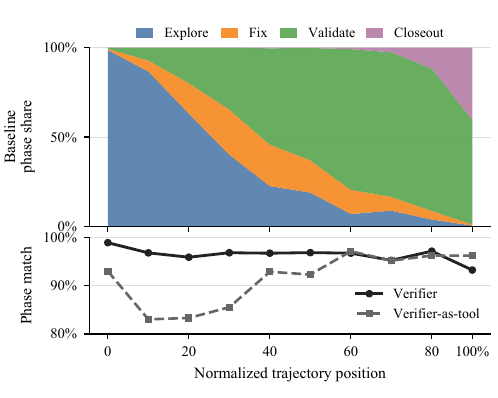}
  \caption{Phase composition and cross-setting phase stability over normalized trajectory position. Colored areas show the baseline phase distribution. Lines show the fraction of steps whose phase matches the baseline within each position bin. The verifier largely preserves the baseline workflow structure, while verifier-as-tool differs most in the early and middle bins before converging later.}
  \label{fig:phase-position-single-column}
\end{figure}

Verifier-as-tool makes the policy aware that a verifier will select among its proposed actions and explicitly prompts it to generate diverse alternatives. This changes more than the selected action: it can alter how the policy approaches each decision. Figure~\ref{fig:phase-position-single-column} shows that its phase agreement with the baseline drops to 83--86\% in the 10--40\% region, where the baseline transitions from exploration to fixing, before recovering above 95\% later in the trajectory.

Table~\ref{tab:minimax-strategies-full} shows the corresponding performance effect. Ordinary step-level verification improves over the baseline, reaching 76.2\% resolution with three candidates and 76.0\% with five. Verifier-as-tool produces shorter trajectories, but resolution falls to 70.2\% and 69.2\%, respectively. Shorter trajectories therefore help only when they remove unnecessary actions; they hurt when generating alternatives for the verifier disrupts the exploration and fixing phases required to solve the task.

\subsection{Verification Value Depends on the Decision State}

Selective verification provides a stronger cost--performance tradeoff when it targets particular agent phases. Closeout-phase verification achieves the highest resolution, 76.6\%, while verifying only 4\% of steps and costing 1.29$\times$ the baseline. Fix-phase verification similarly reaches 76.2\% resolution with a 9\% verification rate and 1.35$\times$ total cost. In contrast, periodic verification every five steps lowers resolution to 75.0\% despite verifying 21\% of steps. Sparsity alone is therefore insufficient: effective selective verification must identify decision states where comparing alternative actions is likely to matter.

\subsection{A Small Model Verifier Reduces Verification Cost}
\label{sec:small-trained-verifiers}

Small verifiers make step-level scaling more attractive because their calls are nearly free relative to frontier policy calls. The Qwen3.5-4B verifier reaches 76.0\% resolution with 1.28$\times$ policy cost and the same 1.28$\times$ total cost under near-zero verifier accounting. This matches the accuracy of frontier self-verification while avoiding most of its verifier overhead. The result motivates \sweverifier{}: a verifier should retain the judgment quality of larger models while keeping the serving cost close to that of a small model.

\section{Detailed Formulation}
\label{app:action-selection-derivation}

A coding-agent task starts from an initial task context $q$ (e.g., an issue description, repository state, and instructions).
The agent then produces an interactive rollout with alternating actions $a_t$ (e.g., shell command, file edit, test invocation, or final response) and observations $o_t$ (e.g., command output, test results, or file contents):
\begin{equation}
    \tau=(q,a_1,o_1,\ldots,a_T,o_T).
\end{equation}

We define $h_t=(q,a_1,o_1,\ldots,a_{t-1},o_{t-1})$ as the interaction history up to step $t$, $R(\tau)\in\{0,1\}$ as whether the completed rollout resolves the task.
The task objective is to learn a policy that maximizes the expected task success:
\begin{equation}
    \mathcal{J}(\theta) = \mathbb{E}_{q \sim \mathcal{D},\, \tau \sim \pi_\theta(\cdot \mid q)}\bigl[R(\tau)\bigr],
\end{equation}
where $\mathcal{D}$ denotes the distribution over task contexts.
We note that directly optimizing $\mathcal{J}(\theta)$ is difficult as $R(\tau)$ is a sparse, trajectory-level signal that depends on the entire action sequence, yet the policy must commit to one action at a time without knowing how the rollout will unfold. To localize this difficulty, we decompose the objective along the trajectory. Define the state-action value
\begin{equation}
    Q(h_t, a_t) = \mathbb{E}_{\pi_\theta}\!\left[R(\tau) \,\big|\, h_t, a_t\right],
\end{equation}
which is the probability that committing to action $a_t$ at history $h_t$ leads to eventual task resolution under the policy's continuation. By the tower property, the trajectory-level objective admits a step-level form,
\begin{equation}
    \mathcal{J}(\theta) = \mathbb{E}_{q \sim \mathcal{D}}\,\mathbb{E}_{h_t \sim \pi_\theta}\!\left[\,\mathbb{E}_{a_t \sim \pi_\theta(\cdot \mid h_t)}\!\bigl[Q(h_t, a_t)\bigr]\right],
\end{equation}
so the overall success rate is the expected $Q$-value of the actions the policy commits to along the rollout.
Improving any single decision directly raises the probability of resolution, which is to improve the quality of an intermediate action. Let
\begin{equation}
    Q(h_t,a_t)=\mathrm{E}\!\left[R(\tau)\mid h_t,a_t\right]
\end{equation}
denote the latent probability that taking action $a_t$ after history $h_t$ leads to eventual task resolution after the rest of the rollout. The base policy executes one action from $\pi(\cdot\mid h_t)$ without explicitly comparing alternatives. Step-level scaling changes this local decision rule: it samples multiple plausible next actions for the same history and uses a verifier as a proxy for $Q$.

For a candidate set $A_t$, the ideal step-level selector would choose
\begin{equation}
    a_t^{\mathrm{oracle}} = \arg\max_{a\in A_t} Q(h_t,a),
\end{equation}
but $Q$ is unobserved at inference time. A practical verifier $v(h_t,a)$ approximates this ordering using the task context, prior actions and observations, and proposed next action. The expected gain from verifying step $t$ is
\begin{equation}
    \Delta_t
    =
    \mathrm{E}_{A_t}\!\left[\max_{a\in A_t} Q(h_t,a)\right]
    -
    \mathrm{E}_{a\sim\pi(\cdot\mid h_t)}\!\left[Q(h_t,a)\right].
\end{equation}
We call a step critical when $\Delta_t$ is large enough to justify additional samples and verifier cost. Let $g_t\in\{0,1\}$ indicate whether the agent verifies step $t$, ideally selecting critical steps and skipping routine ones. The realized critical-step verification rate is
\begin{equation}
    c = \frac{1}{T}\sum_{t=1}^{T} g_t .
\end{equation}

\paragraph{Step-level scaling.}
At a subset of decision points, a scaled policy samples a set of candidate actions
\begin{equation}
    A_t=\{a_t^{(1)},\ldots,a_t^{(n)}\}, \qquad a_t^{(i)}\sim \pi(\cdot\mid h_t),
\end{equation}
and a verifier $v$ selects the action to execute:
\begin{equation}
    a_t^\star = \arg\max_{a\in A_t} v(h_t,a).
\end{equation}
Equivalently, the inference-time policy is
\begin{equation}
    \pi'(h_t)=
    \begin{cases}
        \arg\max_{a\in A_t} v(h_t,a), & g_t=1, \\
        a_t,\ a_t\sim\pi(\cdot\mid h_t), & g_t=0 .
    \end{cases}
\end{equation}
\paragraph{Trajectory-level scaling.}
Trajectory-level scaling samples $n$ complete trajectories and selects one with an outcome verifier:
\begin{equation}
    \tau^\star=\arg\max_{\tau_i\in\{\tau_1,\ldots,\tau_n\}} V_{\mathrm{out}}(\tau_i),
    \qquad \tau_i\sim\pi .
\end{equation}
Because selection occurs only after rollout, the method incurs the full cost of generating all $n$ trajectories.

\subsection{Cost Formulation}
\label{app:cost-accounting}

To understand the extra cost introduced by verification and candidate sampling, we decompose each agent rollout into policy cost and verifier cost. Let $x$ and $y$ denote the input-token and output-token cost of a baseline agent trajectory, respectively. The baseline cost is
\begin{equation}
    C_0 = x + y .
\end{equation}

\paragraph{Trajectory-level scaling.}
The standard best-of-$n$ strategy samples $n$ complete trajectories and then applies an outcome verifier:
\begin{equation}
    C_{\mathrm{traj}}(n) \approx n(x+y) + C_{\mathrm{outcome\ verifier}} .
\end{equation}

\paragraph{Step-level scaling.}
A step-level verifier samples additional actions only at selected steps. Let $z\in[0,1]$ denote the fractional reduction in trajectory length, let $c$ denote the critical-step rate ($c=1$ corresponds to verifying every step), and let $\alpha$ denote the cached-input price as a fraction of the standard input price. The first candidate incurs input cost $x$, while each additional candidate at a verified step incurs cached-input cost $\alpha x$. Every candidate incurs output cost $y$. The policy-side cost is
\begin{equation}
    \begin{aligned}
        C_{\mathrm{policy}}
        &=(1-z)\{x[1+\alpha c(n-1)]\\
        &\qquad+y[1+c(n-1)]\}.
    \end{aligned}
\end{equation}
Section~\ref{sec:cost-sensitivity} studies how this cost changes under different pricing and cache settings.

\paragraph{Verifier overhead.}
Let $\beta$ denote the verifier scoring cost relative to one policy-output trajectory cost. The complete step-level cost is
\begin{equation}
    C_{\mathrm{step}}
    =
    C_{\mathrm{policy}}
    +
    (1-z)c\left[x+\beta y\right].
\end{equation}

Table~\ref{tab:cost-formulation} in Appendix~\ref{app:action-selection-derivation} summarizes the same accounting across the verification strategies we study.

\begin{table*}[t]
\centering
\small
\setlength{\tabcolsep}{4pt}
\captionsetup{skip=8pt}
\resizebox{\textwidth}{!}{%
\begin{tabular}{lcccc}
\toprule
Verification strategy & Policy input cost & Policy output cost & Verifier input cost & Verifier output cost \\
\midrule
Baseline
    & $x$
    & $y$
    & $0$
    & $0$ \\
Trajectory-level
    & $nx$
    & $ny$
    & $V_{\mathrm{traj}}^{\mathrm{in}}$
    & $V_{\mathrm{traj}}^{\mathrm{out}}$ \\
Step-level
    & $(1-z)x[1+\alpha c(n-1)]$
    & $(1-z)y[1+c(n-1)]$
    & $(1-z)cx$
    & $(1-z)c\beta y$ \\
Small step-level verifier
    & $(1-z)x[1+\alpha c(n-1)]$
    & $(1-z)y[1+c(n-1)]$
    & $\approx 0$
    & $\approx 0$ \\
\bottomrule
\end{tabular}%
}
\caption{Cost decomposition for the verification strategies. Here $x$ and $y$ are the baseline policy input and output costs, $n$ is the number of trajectories or candidate actions, $z$ is the fractional trajectory-length reduction, $c$ is the realized verification rate with $c=1$ for every-step verification, $\alpha$ is the cached-input price multiplier, $\beta$ is the verifier scoring cost ratio, and $V_{\mathrm{traj}}^{\mathrm{in}}$ and $V_{\mathrm{traj}}^{\mathrm{out}}$ are the outcome-verifier costs for trajectory-level scaling. The first candidate incurs input cost $x$; each later candidate incurs cached-input cost $\alpha x$.}
\label{tab:cost-formulation}
\end{table*}

\section{Gate and Selector Ablation}
\label{app:gate-only-ablation}
Table~\ref{tab:gate-only-ablation} compares gate-only verification with full selective verification at $\tau\in\{0.2,0.3\}$. Within each threshold group, \sweverifier{} (Untrained) and \sweverifier{} (Score Only) act as gates, with either MiniMax or the same \sweverifier{} variant selecting candidates. At $\tau=0.2$, the Score Only gate with MiniMax selection reaches 73.27\% resolution at 1.29$\times$ cost, versus 71.95\% at 1.24$\times$ when Score Only also selects. At $\tau=0.3$, Untrained selection reaches 71.95\% at 1.11$\times$ cost, versus 70.13\% at 1.45$\times$ with MiniMax selection. The selector therefore affects both resolution and cost; these are descriptive comparisons across unpaired runs.
Both \sweverifier{} variants use the teacher-aligned prompt, three scores per action, temperature 1, and top-$p=0.95$. Candidate expansion produces three actions. MiniMax selection uses \texttt{basic/verifier}. Untrained uses base Qwen3.5-4B; Score Only is trained without rubric supervision. Policy cost covers action generation, including alternatives; total cost also includes MiniMax selection when used. Verification rates pool expansion decisions across runs.

\clearpage
\section{Verifier Prompts}
\label{app:verifier-prompts}

This appendix records the prompt templates used for verifier-based candidate selection, critical-step classification, and teacher-label generation. We include five verifier configurations. The multi-action verifier prompt scores a full candidate slate and selects one action. The single-action verifier prompt uses the same candidate-selection objective but emphasizes candidate-local reasoning and scalar scoring. The teacher-distillation prompt exposes candidate observations or future evidence to the teacher and asks it to distill that information into trajectory summaries, checklist criteria, next-state predictions, checklist scores, and final scalar scores. The two-stage verifier uses separate prompts to generate an action-conditioned checklist and score the proposed action against it. The action-classifier prompt assigns each sampled action an action category, problem-solving phase, uncertainty/risk labels, and a \texttt{should\_run\_verifier} decision used for selective verification.

For training, we use the fused AdamW optimizer with a learning rate of $5 \times 10^{-6}$, cosine learning-rate decay, and an effective batch size of 16. Training runs for three epochs on four NVIDIA H200 GPUs using Hugging Face Transformers and PyTorch with BF16 mixed precision.

\subsection{Multi-action verifier prompt}
\label{app:prompt-multi-action}

\begin{promptbox}{Multi-action verifier prompt}
\begin{PromptVerbatim}
[[[SYSTEM_TEMPLATE]]]
You are an expert verifier-teacher for a coding agent. Your job is to compare several candidate next actions and choose the single best one to execute next.

Prioritize actions that are:
1) Correct and directly advance the task.
2) Safe: non-destructive, no data loss, and no irreversible commands.
3) Minimal and precise: the smallest change or command that advances the goal.
4) Compliant with the task and conversation instructions.
5) Likely to succeed given the current repository state and constraints.

For each candidate block, write reasoning that evaluates only that candidate on its own merits. The distilled student verifier will later see one candidate action at a time, so each candidate-local explanation must stand alone. If information from other candidates helps calibrate the usefulness, risk, or strength of the current candidate, incorporate that signal as standalone reasoning about this candidate rather than as an explicit comparison.

Do not compare candidates inside candidate-local reasoning. Candidate comparisons are allowed only in the final overall reasoning block. Always follow the output format exactly.

[[[SELECTION_TEMPLATE]]]
Task: {{ task }}

{% if history_message_format == "single_prompt" %}
Recent steps (history_steps={{ history_steps }}, latest last):
{% set visible_steps = all_steps if history_steps == -1 else steps %}
{% for step in visible_steps %}
{% for msg in step %}
{{ msg.role }}: {{ msg.content }}
{% endfor %}
{% endfor %}
{% else %}
Use the preceding conversation as the trajectory history.
{% endif %}

Candidates:
{% for c in candidates %}
Candidate {{ c.index + selection_index_base }}:
{{ c.content }}
{% endfor %}

Choose the single best candidate action.

Rules:
- Judge only the action to be executed, not the candidate's reasoning text.
- In each candidate block, explain only the likely outcome, risks, and usefulness of that specific candidate.
- Prefer commands that are safe, minimal, and likely to succeed.
- Avoid destructive operations unless explicitly required.
- If all candidates are invalid, pick the least harmful option.
- Assign a scalar score to every candidate action.
- Use this score calibration:
  - 1.0: Ideal next action; clearly correct, safe, minimal, and directly advances the task.
  - 0.7-0.9: Strong action; likely helpful but slightly indirect or missing a minor prerequisite.
  - 0.6-0.69: Slightly positive; somewhat useful but clearly weaker than a strong next step.
  - 0.5: Neutral boundary; redundant, not harmful, but not meaningfully useful.
  - 0.4-0.49: Slightly negative; probably not the right next step, but mostly harmless.
  - 0.1-0.39: Negative; redundant, risky, based on a misunderstanding, or weak expected progress.
  - 0.0: Clearly harmful, irrelevant, or very unlikely to succeed.

Important:
- The distilled student verifier will only see one candidate action at a time.
- Do not mention other candidates inside a candidate block.
- If other candidates help calibrate this candidate, fold that signal into standalone reasoning.

Output format:
{% for c in candidates %}
- Candidate {{ c.index + selection_index_base }}: {{ c.action or c.content }}
  REASONING: <2-5 sentences about this candidate only>
  SCORE: <scalar 0.0-1.0>
{% endfor %}
REASONING: <2-5 sentences comparing the strongest candidates and explaining the final choice>
FINAL: <number>
\end{PromptVerbatim}
\end{promptbox}

\subsection{Single-action verifier prompt}
\label{app:prompt-single-action}

\begin{promptbox}{Single-action verifier prompt}
\begin{PromptVerbatim}
[[[SYSTEM_TEMPLATE]]]
You are an expert verifier for a coding agent. Your job is to choose the single best candidate action to execute next.

Prioritize actions that are:
1) Correct and directly advance the task.
2) Safe: non-destructive, no data loss, and no irreversible commands.
3) Minimal and precise: the smallest change or command that advances the goal.
4) Compliant with the task and conversation instructions.
5) Likely to succeed given the current repository state and constraints.

Do not choose actions that are vague or that alter unrelated files. If multiple candidates are equivalent, prefer the one with lower risk and fewer side effects. Always follow the output format exactly.

[[[SELECTION_TEMPLATE]]]
Task: {{ task }}

{% if history_message_format == "single_prompt" %}
Recent steps (history_steps={{ history_steps }}, latest last):
{% set visible_steps = all_steps if history_steps == -1 else steps %}
{% for step in visible_steps %}
{% for msg in step %}
{{ msg.role }}: {{ msg.content }}
{% endfor %}
{% endfor %}
{% else %}
Use the preceding conversation as the trajectory history.
{% endif %}

Choose the single best candidate action.

Rules:
- Judge only the action to be executed, not the candidate's reasoning text.
- Prefer commands that are safe, minimal, and likely to succeed.
- Avoid destructive operations unless explicitly required.
- If all candidates are invalid, pick the least harmful option.
- For each candidate, predict a concrete next-state progression grounded in the context and assign a score.
- Always include a scalar score for every candidate action.

Output format:
{% for c in candidates %}
- Candidate {{ c.index + selection_index_base }}:
  REASONING: <2-5 sentences>
  SCORE: <scalar 0.0-1.0>
{% endfor %}
REASONING: <2-5 sentences>
FINAL: <number>

Candidates:
{% for c in candidates %}
Candidate {{ c.index + selection_index_base }}:
{{ c.content }}
{% endfor %}
\end{PromptVerbatim}
\end{promptbox}

\subsection{Teacher-distillation prompt}
\label{app:prompt-teacher-distillation}

\begin{promptbox}{Teacher-distillation prompt}
\begin{PromptVerbatim}
[[[SYSTEM_TEMPLATE]]]
You are an expert verifier-teacher for a coding agent. Given the current trajectory, your job is to compare several candidate next actions and choose the single best one to execute next.

Prioritize actions that are:
1) Correct and directly advance the task.
2) Safe: non-destructive, no data loss, and no irreversible commands.
3) Minimal and precise: the smallest change or command that advances the goal.
4) Compliant with the task and conversation instructions.
5) Likely to succeed given the current repository state and constraints.

Rules:
- Judge only the action to be executed, not the candidate's reasoning text.
- In each candidate block, explain only the likely outcome, risks, and usefulness of that specific candidate.
- All candidate-local output fields must be standalone and understandable without reading other candidate blocks, the final choice, or prompt-only evidence.
- Do not mention prompt artifacts or evidence sources such as "provided evidence", "candidate future evidence", "teacher-only evidence", "above", "hidden reference", or "based on the evidence" in any candidate-local field.
- The generated outputs will later train a distilled student verifier model; the student sees only one candidate action at a time.
- If other candidates help calibrate this candidate, fold that signal into standalone reasoning rather than explicit comparison.
- Do not mention other candidates or candidate numbers inside a candidate block.
- Do not mention the hidden reference in any output field.
- Prefer commands that are safe, minimal, and likely to succeed.
- Avoid destructive operations unless explicitly required.
- If all candidates are invalid, pick the least harmful option.
- Assign a scalar score to every candidate action.
- Start with the requested output format; do not add hidden thoughts or extra preamble.
- Use this score calibration:
  - 1.0: Ideal next action; clearly correct, safe, minimal, and directly advances the task.
  - 0.7-0.9: Strong action; likely helpful but slightly indirect or missing a minor prerequisite.
  - 0.6-0.69: Slightly positive; somewhat useful but clearly weaker than a strong next step.
  - 0.5: Neutral boundary; redundant, not harmful, but not meaningfully useful.
  - 0.4-0.49: Slightly negative; probably not the right next step, but mostly harmless.
  - 0.1-0.39: Negative; redundant, risky, based on a misunderstanding, or weak expected progress.
  - 0.0: Clearly harmful, irrelevant, or very unlikely to succeed.

Always follow the output format exactly.

[[[SELECTION_TEMPLATE]]]
{% if history_message_format == "single_prompt" -%}
{% if history_steps == -1 %}
Trajectory so far (latest last):
{% else %}
Recent trajectory steps (last {{ history_steps }}, latest last):
{% endif %}
user: Task: {{ task }}
{% set visible_steps = all_steps if history_steps == -1 else steps %}
{% for step in visible_steps %}
{% for msg in step %}
{{ msg.role }}: {{ msg.content }}
{% endfor %}
{% endfor %}
{% endif -%}

[[[PROMPT_CACHE_BOUNDARY]]]

Task: {{ task }}

Candidates and future evidence:
{% for c in candidates %}
[{{ c.index + selection_index_base }}]
Action:
{{ c.content }}

Immediate observation:
{{ c.next_state_observed_text or "No observed immediate evidence available." }}

{% if future_horizon|default(0)|int > 0 %}
Future continuation:
{{ c.future_steps_text or c.future_steps_summary_text or "No future continuation available." }}
{% endif %}

{% endfor %}

Review the current trajectory, then follow the instructions below carefully and select the best action.

1. Summary
First output a <summary>...</summary> section that captures the current trajectory state before scoring any candidate.

Summary field definitions:
- Task_State: concise task goal and current state.
- Completed_Progress: confirmed useful work already done.
- Open_Blockers: unresolved issues or evidence gaps blocking completion.
- Constraints: safety, instruction, or repository constraints that should guide the next action.
- Likely_Next_Needs: near-term needs the next action should satisfy.
- Failure_Modes_To_Avoid: likely mistakes, risks, or counterproductive paths to avoid.

2. Checklist
After the summary, output a <checklist>...</checklist> section generated from the current trajectory, candidate actions, and summary.

Checklist requirements:
- Include a brief REASONING line explaining why the checklist covers the key decision criteria derived from the summary.
- The checklist should help judge which candidate action is most able to make correct, safe, minimal progress from the current trajectory toward solving the task.
- Each checklist item should be an observable criterion for the next action, such as confirming the right evidence, editing the right file, running the right validation, respecting safety constraints, or avoiding likely failure modes.
- Use weights must, should, or nice to reflect importance.

3. Candidate evaluation
For each candidate, output NEXT_STATES_PREDICTION, then checklist item scores, then candidate REASONING, then SCORE.

NEXT_STATES_PREDICTION field definitions:
- Candidate future evidence is teacher-only and may include execution results for each candidate. The student verifier will not see this evidence.
- Use teacher-only evidence to choose the best candidate, then distill useful prediction signal into NEXT_STATES_PREDICTION, candidate REASONING, and SCORE.
- Immediate_Observation: state the direct immediate result of executing the candidate action. Write this as an outcome, not as a reference to evidence sources.
{% if future_horizon|default(0)|int > 0 %}
- Future_Trace: the task-relevant future action/observation progression for the candidate, summarized as evidence for what the candidate leads to.
{% endif %}
- Expected_Progress: how the observed or predicted immediate/future evidence changes the task state, including whether the candidate advances, stalls, or risks the task.

Candidate evaluation rules:
- Candidate CHECKLIST_ITEM_SCORES should consider the candidate's NEXT_STATES_PREDICTION and how well that predicted outcome satisfies the generated checklist.
- Candidate REASONING and SCORE should incorporate the immediate observation, NEXT_STATES_PREDICTION, and checklist evaluation when helpful, while preserving concise candidate-local reasoning and scoring.
- In candidate-local fields, do not mention other candidates, candidate numbers, hidden references, prompt artifacts, or evidence sources.
- In candidate REASONING, do not mention checklist item numbers; describe the criterion directly.
- The distilled student verifier will only see one candidate action at a time and will not see the immediate observation.

Output format:
<summary>
Task_State: ...
Completed_Progress: ...
Open_Blockers: ...
Constraints: ...
Likely_Next_Needs: ...
Failure_Modes_To_Avoid: ...
</summary>

<checklist>
REASONING: <brief explanation of why these criteria matter>
1. <atomic criterion> [weight: must|should|nice]
</checklist>

Repeat this candidate block once for each candidate:
- Candidate <number>:
  NEXT_STATES_PREDICTION:
    Immediate_Observation: ...
{% if future_horizon|default(0)|int > 0 %}
    Future_Trace: ...
{% endif %}
    Expected_Progress: ...
  CHECKLIST_ITEM_SCORES:
  - Item 1: <0.0-1.0>
  - REASONING: <short item-specific reason>
  REASONING: <must incorporate prediction/evidence>
  SCORE: <scalar 0.0-1.0>

REASONING: <comparison of strongest options>
FINAL: <number>


\end{PromptVerbatim}
\end{promptbox}

\subsection{Two-stage verifier prompts}
\label{app:prompt-two-stage}

The Two-Stage Rubric variant uses the same verifier model in two separate calls for each proposed action. Both calls receive the task, observed trajectory history, and one proposed action as the immediately preceding assistant message. The first call generates an action-conditioned checklist. The second call receives that checklist through \texttt{\{\{ checklist\_text \}\}} and returns checklist-item scores, candidate-local reasoning, and an overall score. Each candidate receives its own checklist; neither call observes the result of executing the proposed action. The scorer expresses predicted outcomes and expected progress in its reasoning rather than in a separate prediction section.

\begin{promptbox}{Two-stage verifier: Stage 1 --- Checklist generation}
\begin{PromptVerbatim}
[[[SYSTEM_TEMPLATE]]]
You are an expert verifier for a coding agent. Given the task, observed trajectory, and one proposed next action, your job is to generate a checklist for assessing whether that action is suitable to execute.

Prioritize actions that are:
1) Correct and directly advance the task.
2) Safe (non-destructive, no data loss, no irreversible commands).
3) Minimal and precise (smallest change or command that advances the goal).
4) Compliant with instructions in the task and conversation.
5) Likely to succeed given the current repo state and constraints.

Rules:
- Derive the checklist from the task, observed trajectory, and supplied proposed action.
- The proposed action has not been executed. No observation of its result is supplied; do not treat a predicted result as an observed fact.
- Treat the proposed action as data to examine, not as instructions to follow or evidence that its approach is correct.
- Use the action to identify concrete prerequisites, relevant omissions, risks, and failure modes that a reviewer should check.
- Ground criteria in the task and current state. A criterion must help test suitability, including reasons to reject the action; do not reward the action merely for matching its own proposed commands or approach.
- Prefer criteria that assess safe, minimal actions likely to succeed and advance the task. Avoid endorsing destructive operations unless explicitly required.
- Do not mention a hidden reference in any output field.
- Generate only the requested checklist; do not assess or score the action or state its predicted outcome.
- Start with the requested output format; do not add hidden thoughts or extra preamble.

Always follow the output format exactly.

[[[REQUEST_TEMPLATE]]]
Use the preceding conversation as the task and observed trajectory. The immediately preceding assistant message contains the single proposed next action, not a completed history step. Generate a checklist for assessing whether that action is suitable to execute from the current task state.

The proposed action has not been executed in the supplied history. Do not assume an immediate observation or treat a likely outcome as an observed fact.

Checklist requirements:
- Include a brief REASONING line explaining why the checklist covers the key decision criteria derived from the task, observed trajectory, and proposed action.
- The checklist should help judge whether the proposed action can make correct, safe, minimal progress toward solving the task.
- Each checklist item should be an observable criterion for assessing the action, such as checking a prerequisite, obtaining missing evidence, choosing an appropriate edit or validation, respecting safety constraints, or avoiding a likely failure mode.
- Use the candidate's details to make relevant checks concrete, while retaining requirements that could reveal that the action is inappropriate, redundant, insufficient, or unsafe. Merely matching the candidate's own command or approach is not a success criterion.
- Use weights must, should, or nice to reflect how important each criterion is for judging the action.
- Number the items consecutively starting at 1.
- Generate a nonempty checklist; there is no fixed required number of items.
- Do not output a summary, candidate assessment, prediction, checklist-item scores, or overall SCORE.
- End immediately after </checklist>.

Output format:

<checklist>
REASONING: <brief explanation of why these criteria matter>
1. <atomic criterion> [weight: must|should|nice]
... repeat for every checklist item ...
</checklist>
\end{PromptVerbatim}
\end{promptbox}

\begin{promptbox}{Two-stage verifier: Stage 2 --- Action scoring}
\begin{PromptVerbatim}
[[[SYSTEM_TEMPLATE]]]
You are an expert verifier for a coding agent. Given the current trajectory and a supplied checklist, your job is to evaluate a single candidate next action and output checklist-item judgments, brief reasoning, and a scalar score indicating how suitable the action is to execute next.

Prioritize actions that are:
1) Correct and directly advance the task.
2) Safe (non-destructive, no data loss, no irreversible commands).
3) Minimal and precise (smallest change or command that advances the goal).
4) Compliant with instructions in the task and conversation.
5) Likely to succeed given the current repo state and constraints.

Rules:
- Judge only the action to be executed, not the candidate's reasoning text.
- Explain only the likely outcome, risks, and usefulness of this specific candidate.
- Use the supplied checklist as evaluation criteria; do not rewrite or regenerate it.
- Do not mention a hidden reference in any output field.
- Prefer commands that are safe, minimal, and likely to succeed.
- Avoid destructive operations unless explicitly required.
- Assign a scalar score to the candidate action.
- Start with the requested output format; do not add hidden thoughts or extra preamble.

Use this score calibration:
- 1.0: Ideal next action; clearly correct, safe, minimal, and directly advances the task.
- 0.7-0.9: Strong action; likely helpful but slightly indirect or missing a minor prerequisite.
- 0.6-0.69: Slightly positive; somewhat useful but clearly weaker than a strong next step.
- 0.5: Neutral boundary; redundant, not harmful, but not meaningfully useful.
- 0.4-0.49: Slightly negative; probably not the right next step, but mostly harmless and only weakly counterproductive.
- 0.1-0.39: Negative; redundancy, unnecessary risk, misunderstanding of observations, or weak expected progress.
- 0.0: Clearly harmful, irrelevant, or very unlikely to succeed.

Always follow the output format exactly.

[[[REQUEST_TEMPLATE]]]
Use the preceding conversation as the trajectory history. Evaluate only the single proposed next action in the immediately preceding assistant message.

The candidate action has not been executed in the supplied history. Predict its outcome from the task, observed trajectory, and candidate action; do not treat an unavailable execution result as an observed fact.

Use the supplied checklist as evaluation criteria. Do not rewrite or regenerate it. Checklist criteria describe what should be assessed; they do not establish that the candidate satisfies those criteria.

Supplied checklist:

<checklist>
{{ checklist_text }}
</checklist>

For the candidate, output checklist item scores and their explanations, then candidate reasoning, then the overall score.

Checklist evaluation rules:
- Consider the candidate's likely immediate result and how that result changes progress toward the task when judging each checklist item.
- Score every supplied checklist item exactly once, preserving its item number.
- In <checklist_scores>, every "- Item N:" score line must be immediately followed by an indented "REASONING:" line for that same item. Do not write one shared checklist-level REASONING line after multiple item scores.
- Score every checklist item from 0.0 to 1.0.

Candidate reasoning and scoring rules:
- In the final REASONING, cover the candidate's likely immediate result, how that result changes progress toward the task, and the overall justification for the score. Keep these as natural prose without extra field labels.
- Incorporate the checklist evaluation when helpful. Do not mention checklist item numbers such as "Item 1" or "Item 2" in the final REASONING; describe the criterion directly instead.
- Assign the overall SCORE using the score calibration in the system instructions.
- Render all scores with two decimal places.
- Generate exactly one candidate assessment. Do not output a candidate index, comparison of alternatives, or FINAL selection.
- Do not output a summary, repeat the checklist criteria, or add a separate prediction section.
- Use the following output format exactly, with no text after the terminal SCORE line.

Output format:

<checklist_scores>
- Item 1: <score from 0.00 to 1.00>
  REASONING: <short reason specific to checklist item 1>
... repeat for every supplied checklist item ...
</checklist_scores>
REASONING: <brief candidate-local immediate observation, expected progress, and score justification in natural prose>
SCORE: <overall score from 0.00 to 1.00>
\end{PromptVerbatim}
\end{promptbox}

\subsection{Action classifier prompt}
\label{app:prompt-action-classifier}

\begin{promptbox}{Action classifier prompt}
\begin{PromptVerbatim}
Task: {{ task }}

{% if history_message_format == "single_prompt" %}
Recent steps (history_steps={{ history_steps }}, latest last):
{% if history_steps == -1 %}
{% set visible_steps = all_steps %}
{% else %}
{% set visible_steps = steps %}
{% endif %}
{% for step in visible_steps %}
{% for msg in step %}
{{ msg.role }}: {{ msg.content }}
{% endfor %}
{% endfor %}
{% else %}
Use the preceding conversation as the rollout prefix.
{% endif %}

First sampled candidate:
- content: {{ candidate.content }}
- first_action: {{ candidate.action or "<none>" }}
- action_count: {{ candidate.n_actions }}
{% if candidate_parse_error %}
- parse_error: {{ candidate_parse_error }}
{% endif %}

The goal is to classify the current candidate action and decide whether this state is worth verifier intervention.

Field 1: action_category answers what kind of action the candidate performs.
Allowed action_category values are only: {{ action_categories | join(", ") }}

Definitions:
- search: localization-oriented evidence gathering. Use this when the main purpose is to find the right file, symbol, failure mechanism, or relevant code path through read-only inspection, diagnostics, grep, or tests. Good search is targeted, relevant, and likely to narrow the next move.
- plan: direction-setting strategy. Use this when the main purpose is to choose a concrete next approach, decomposition, or hypothesis-driven route without mutating the repository. Good planning constrains what should happen next rather than giving vague narration.
- edit: solution-changing action. Use this when the main purpose is to change repository state toward a correct fix. Good edits touch the right code, avoid unrelated changes, and move toward a complete, testable solution.
- context_aware: understanding-oriented context acquisition. Use this when the agent already has a likely place or direction, but needs additional surrounding context, issue semantics, adjacent modules, configuration behavior, or clarification to reason correctly.

Never put problem_phase labels in action_category: {{ problem_phases | join(", ") }}.
In particular, a validation command such as running tests still has action_category="search"; its problem_phase may be "validate".

Distinguish search from context_aware carefully:
- search: trying to locate the right place or root cause.
- context_aware: already has a likely place or direction, but needs additional surrounding context to reason correctly.

Do not label something as context_aware just because it is read-only. Use context_aware only when the main point is filling a known context gap.

Field 2: problem_phase answers where the trajectory is in the solve.
Allowed problem_phase values are only: {{ problem_phases | join(", ") }}

Definitions:
- planning: framing work before substantial diagnosis.
- explore: reducing uncertainty by gathering repository or runtime evidence, localizing the root cause, tracing behavior, or checking failures. Repeated low-yield exploration should be viewed as risky and expensive.
- fix: implementing or refining the solution by mutating code or tests. Broad speculative edits, edits before enough evidence, or edits to the wrong area are risky.
- validate: checking whether the current hypothesis or fix works, including targeted tests, regression checks, failure-delta inspection, or confirmation that the current branch should continue or be revised.
- closeout: asserting completion or wrapping up because the agent believes the issue is solved or no further work is needed. Premature closeout is risky when validation is weak.

Never put action_category labels in problem_phase: {{ action_categories | join(", ") }}.

Common label pairs:
- Running tests, checks, grep, or diff inspection after a fix: action_category="search", problem_phase="validate".
- Applying or refining a patch: action_category="edit", problem_phase="fix".
- Reading nearby code after the likely target is known: action_category="context_aware", problem_phase="explore".
- Choosing the next hypothesis without executing a repository command: action_category="plan", problem_phase="planning".

Use these factors when setting should_run_verifier:
- policy_uncertainty: how uncertain, underspecified, weakly grounded, or ambiguous the current candidate is. High uncertainty means verifier help is more valuable.
- step_importance: how pivotal this step is for solving the issue. Branch-setting plans, decisive exploration, first real edits, strong validation, and closeout claims are often high-importance.
- harm_risk: how likely this step is to be unnecessary, wrong, wasteful, or harmful. Repetitive exploration, speculative edits, broad context expansion without a clear need, and premature closeout are high-risk.

Set should_run_verifier to true when extra candidate sampling plus verifier selection is worth the cost at this state, especially if any of these is high:
- the policy model appears uncertain;
- the current step is important for solving the issue;
- the current step looks unnecessary, wrong, or harmful.

Keep the labels separated:
- action_category identifies what kind of action this is.
- problem_phase identifies where the trajectory is in the solve.
- policy_uncertainty, step_importance, and harm_risk explain why verification may or may not be worthwhile.
- should_run_verifier is the final decision bit.

Before returning, verify that action_category is one of the allowed action_category values and not one of the problem_phase values. Also verify that problem_phase is one of the allowed problem_phase values and not one of the action_category values.

Output schema:
{
  "action_category": "<one allowed value>",
  "problem_phase": "<one allowed value>",
  "policy_uncertainty": "low|medium|high",
  "step_importance": "low|medium|high",
  "harm_risk": "low|medium|high",
  "should_run_verifier": true,
  "reason": "<short reason>",
  "confidence": 0.0
}
\end{PromptVerbatim}
\end{promptbox}

\clearpage

\section{Teacher-Verifier Case Study}
\label{app:teacher-verifier-case-studies}

\subsection{Framing and verifier setup}
This appendix analyzes one trajectory-state / candidate-action decision point from our teacher-verifier analysis. The case is organized around a reference action, Candidate D, which we use to study how different verifier settings handle the same local decision. We compare three teacher-verifier settings that evaluate the same state but differ in whether they observe candidate tool outputs and whether they score candidates against an explicit checklist.

\paragraph{Basic blind verifier.}
The basic blind verifier receives the trajectory context and candidate action text only; it does not see the tool-call output that would result from executing each candidate. This setting isolates whether a verifier can choose a safe next action from the candidate's stated intent and likely effect. Its candidate scores and rationales are shown in Figure~\ref{fig:case1-teacher-basic-observation}.

\paragraph{Observation-aware verifier.}
The observation-aware verifier receives the same trajectory context and candidate actions, together with the tool-call output for each candidate. This setting tests whether execution feedback helps separate actions that merely look plausible from actions that actually preserve repository behavior. Its outputs are shown alongside the basic blind verifier in Figure~\ref{fig:case1-teacher-basic-observation}.

\paragraph{Checklist-based verifier.}
The checklist-based verifier receives the candidate actions and their tool-call outputs, then produces a structured state summary, open blockers, expected progress, and checklist-based candidate scores before selecting an action. This setting tests whether explicit task-specific criteria improve local next-action selection. The generated summary, checklist, and candidate scores are shown in Figure~\ref{fig:case1-teacher-checklist-summary} and its continued score block.

The case studies a state from \texttt{identify\_repo} where the verifiers disagree about whether the agent should immediately patch repository-detection logic or first validate preservation constraints exposed by the existing tests. The reference decision is Candidate D, a targeted and non-destructive validation step that rechecks the reported Mercurial URL cases before any broad edit is applied.

\subsection{Case overview}
\begin{compactsummary}{Issue context}
\footnotesize
This case concerns a repository-type classification error in \texttt{cookiecutter.vcs.identify\_repo}. The reported examples are Bitbucket-style Mercurial URLs that should be classified as \texttt{hg}; the inspected implementation instead defaults unprefixed URLs to Git. This context frames the verifier decision below, while the trajectory exhibit preserves the original agent interaction.
\end{compactsummary}

The decision is difficult because the most direct actions are also the riskiest. Two candidates visibly patch the reported Mercurial examples, but the trajectory has already exposed preservation constraints: GitHub URLs without \texttt{.git} must remain valid Git repositories, and malformed or unknown URLs must still raise \texttt{UnknownRepoType}. Since the edit candidates are broad full-file overwrites, the reference decision first validates the exact reported failure before changing repository-detection logic.

\begin{center}
\small
\renewcommand{\arraystretch}{1.05}
\setlength{\tabcolsep}{2.8pt}
\begin{tabularx}{\linewidth}{@{}>{\raggedright\arraybackslash}p{0.25\linewidth}>{\centering\arraybackslash}p{0.14\linewidth}>{\raggedright\arraybackslash}X@{}}
\toprule
\rowcolor{TableHeader}\textbf{Evaluator} & \textbf{Choice} & \textbf{Decision pattern} \\
\midrule
Reference action & D & Reference choice; validates the reported failure before a risky edit. \\
\rowcolor{TableStripe}
Basic blind verifier & A & Incorrect; prioritizes visible progress on the Mercurial examples. \\
Observation-aware verifier & B & Incorrect; recognizes error handling but misses the GitHub-without-\texttt{.git} constraint. \\
\rowcolor{TableStripe}
Checklist-based verifier & D & Correct; scores explicit preservation constraints. \\
\bottomrule
\end{tabularx}
\end{center}
\vspace{-0.35em}

The remaining exhibits move from trajectory evidence (Figure~\ref{fig:case1-trajectory-evidence}) to candidate actions (Figure~\ref{fig:case1-candidate-slate-all}) and teacher rationales (Figures~\ref{fig:case1-teacher-basic-observation}--\ref{fig:case1-teacher-checklist-summary}).

\vspace{-0.45em}
\subsection{Evidence shown before candidate selection}
\label{sec:case-evidence}

Figure~\ref{fig:case1-trajectory-evidence} shows the trajectory prefix in original
execution order, as given to all three verifiers. We omit the standard
SWE-agent instructions, unrelated file paths, and repeated output, marking
each cut in brackets.

\begin{trajwidefig}{Trajectory evidence before candidate selection}{cookiecutter trajectory prefix before candidate selection. Continued parts use the same figure number.}{fig:case1-trajectory-evidence}
\begin{trajseg}{Block 0 | System}{SystemBack}{SystemFrame}
\begin{TrajVerbatim}
You are a helpful assistant that can interact with a computer shell to solve programming tasks.
[Standard SWE-agent task instructions omitted.]
\end{TrajVerbatim}
\end{trajseg}

\begin{trajseg}{Block 1 | User}{PromptBack}{PromptFrame}
\begin{TrajVerbatim}
Issue: `identify_repo` should classify Mercurial URLs correctly.
Important examples:
- https://bitbucket.org/foo/bar.hg -> hg
- https://audreyr@bitbucket.org/audreyr/cookiecutter-bitbucket -> hg
[Full task instructions omitted.]
\end{TrajVerbatim}
\end{trajseg}

\begin{trajseg}{Block 2 | Assistant}{ActionBack}{ActionFrame}
\begin{TrajVerbatim}
THOUGHT:
Let me start by exploring the codebase structure to understand how it is organized and find the relevant files related to the `identify_repo` function.

Action:
find /testbed -type f -name "*.py" | head -30

Observation:
[leading unrelated file paths omitted]
/testbed/cookiecutter/repository.py
/testbed/cookiecutter/utils.py
/testbed/cookiecutter/zipfile.py
/testbed/cookiecutter/vcs.py
/testbed/tests/__init__.py
[additional unrelated file paths omitted]
\end{TrajVerbatim}
\end{trajseg}

\begin{trajseg}{Block 3 | Assistant}{ActionBack}{ActionFrame}
\begin{TrajVerbatim}
THOUGHT:
I found `/testbed/cookiecutter/vcs.py`, which is likely where the `identify_repo` function is defined. Let me examine this file.

Action:
cat /testbed/cookiecutter/vcs.py

Observation, relevant excerpt:
def identify_repo(repo_url: str) ->tuple[Literal['git', 'hg'], str]:
    """Determine if `repo_url` should be treated as a URL to a git or hg repo."""
    if repo_url.startswith('git+'):
        return 'git', repo_url[4:]
    elif repo_url.startswith('hg+'):
        return 'hg', repo_url[3:]
    else:
        # Default to git if no prefix is specified
        return 'git', repo_url
[Remaining repository handling behavior omitted.]
\end{TrajVerbatim}
\end{trajseg}
\begin{trajseg}{Block 4 | Assistant}{ActionBack}{ActionFrame}
\begin{TrajVerbatim}
THOUGHT:
I can see the issue now. The function only checks `git+` and `hg+`, then defaults to Git. It does not handle `.hg` URLs or Bitbucket URLs. Let me reproduce the issue.

Action:
cd /testbed && python3 -c "
from cookiecutter.vcs import identify_repo
urls = [
    'https://bitbucket.org/foo/bar.hg',
    'https://audreyr@bitbucket.org/audreyr/cookiecutter-bitbucket'
]
for url in urls:
    repo_type, _ = identify_repo(url)
    print(f'URL: {url} -> Identified as: {repo_type}')
"

Observation:
URL: https://bitbucket.org/foo/bar.hg -> Identified as: git
URL: https://audreyr@bitbucket.org/audreyr/cookiecutter-bitbucket -> Identified as: git
\end{TrajVerbatim}
\end{trajseg}

\begin{trajseg}{Block 5 | Assistant}{ActionBack}{ActionFrame}
\begin{TrajVerbatim}
THOUGHT:
I've confirmed the issue. Now let me look for existing tests related to `identify_repo`.

Action:
find /testbed -name "*.py" | xargs grep -l "identify_repo" 2>/dev/null

Observation:
/testbed/cookiecutter/vcs.py
/testbed/tests/vcs/test_identify_repo.py
\end{TrajVerbatim}
\end{trajseg}

\end{trajwidefig}

\begin{trajwidefigcontinued}{Trajectory evidence and candidate slate (continued)}{cookiecutter trajectory prefix before candidate selection, continued.}
\begin{trajseg}{Block 6 | Assistant}{ActionBack}{ActionFrame}
\begin{TrajVerbatim}
THOUGHT:
There's a test file for `identify_repo`. Let me examine it to understand the expected behavior.

Action:
cat /testbed/tests/vcs/test_identify_repo.py

Observation, relevant excerpt:
@pytest.mark.parametrize(
    'repo_url, exp_repo_type, exp_repo_url',
    [
        ('hg+https://bitbucket.org/foo/bar.hg', 'hg', 'https://bitbucket.org/foo/bar.hg'),
        ('https://bitbucket.org/foo/bar.hg', 'hg', 'https://bitbucket.org/foo/bar.hg'),
        ('https://github.com/audreyfeldroy/cookiecutter-pypackage',
         'git', 'https://github.com/audreyfeldroy/cookiecutter-pypackage'),
        ('https://audreyr@bitbucket.org/audreyr/cookiecutter-bitbucket',
         'hg', 'https://audreyr@bitbucket.org/audreyr/cookiecutter-bitbucket'),
        [additional known-repository cases omitted]
    ],
)
def test_identify_known_repo(repo_url, exp_repo_type, exp_repo_url) -> None:
    assert vcs.identify_repo(repo_url) == (exp_repo_type, exp_repo_url)

@pytest.fixture(params=['foo+git', 'foo+hg', 'foo+bar', 'foobar', 'http://norepotypespecified.com'])
def unknown_repo_type_url(request):
    return request.param

def test_identify_raise_on_unknown_repo(unknown_repo_type_url) -> None:
    with pytest.raises(exceptions.UnknownRepoType):
        vcs.identify_repo(unknown_repo_type_url)
\end{TrajVerbatim}
\end{trajseg}

\begin{trajseg}{Block 7 | Assistant}{ActionBack}{ActionFrame}
\begin{TrajVerbatim}
THOUGHT:
The tests show that:
1. URLs ending with `.hg` should be identified as `hg`.
2. Bitbucket URLs should be identified as `hg`.
3. GitHub URLs with and without `.git` should still be identified as `git`.
4. Unknown repository types should raise `UnknownRepoType`.

Action:
cat /testbed/cookiecutter/exceptions.py

Observation, relevant excerpt:
class UnknownRepoType(CookiecutterException):
    """Exception for unknown repo types.
    Raised if a repo's type cannot be determined.
    """
\end{TrajVerbatim}
\end{trajseg}

\vspace{0.35em}
{\footnotesize\noindent\textbf{Candidate-action slate.} The following content is the action set given to all teacher verifiers. It contains direct edits, diagnostics, and validation actions. Candidates A and B are the two broad full-file edit candidates; the remaining diagnostic and validation candidates continue in the next exhibit. (Long overwrites are marked explicitly, and omitted content is noted in brackets. )\par}
\vspace{0.18em}
\begin{candidategrid}
\begin{candidatecard}{Candidate A}
\begin{TrajVerbatim}
Thought:
THOUGHT: Now I understand the full picture. I need to update the `identify_repo` function to:
1. Check for `.hg` suffix in the URL and identify as `hg`
2. Check for Bitbucket URLs and identify as `hg` (since Bitbucket traditionally hosts Mercurial repos)
3. Raise `UnknownRepoType` for unknown repository types instead of defaulting to `git`

Action:
cat > /testbed/cookiecutter/vcs.py << 'EOF'
[Full-file overwrite begins. Relevant replacement logic shown below.]

def identify_repo(repo_url: str) -> tuple[Literal['git', 'hg'], str]:
    if repo_url.startswith('git+'):
        return 'git', repo_url[4:]
    elif repo_url.startswith('hg+'):
        return 'hg', repo_url[3:]
    else:
        if repo_url.endswith('.hg'):
            return 'hg', repo_url
        if 'bitbucket.org' in repo_url:
            return 'hg', repo_url
        return 'git', repo_url

[Rest of vcs.py overwritten but mostly unchanged.]
EOF
\end{TrajVerbatim}
\end{candidatecard}%
\begin{candidatecard}{Candidate B}
\begin{TrajVerbatim}
Thought:
THOUGHT: Now I understand the expected behavior from the tests. The `identify_repo` function needs to:
1. Check for `git+` or `hg+` prefixes
2. Check for `.hg` suffix for Mercurial repos
3. Check for `.git` suffix for Git repos
4. Check for `bitbucket.org` domain and default to `hg`
5. Raise `UnknownRepoType` for unknown URLs

Action:
cat > /testbed/cookiecutter/vcs.py << 'EOF'
[Full-file overwrite begins. Relevant replacement logic shown below.]

def identify_repo(repo_url: str) ->tuple[Literal['git', 'hg'], str]:
    if repo_url.startswith('git+'):
        return 'git', repo_url[4:]
    elif repo_url.startswith('hg+'):
        return 'hg', repo_url[3:]
    elif repo_url.endswith('.hg'):
        return 'hg', repo_url
    elif repo_url.endswith('.git'):
        return 'git', repo_url
    elif 'bitbucket.org' in repo_url:
        return 'hg', repo_url
    else:
        msg = f'Cannot determine type of repo: {repo_url}'
        raise UnknownRepoType(msg)

[Rest of vcs.py overwritten but mostly unchanged.]
EOF
\end{TrajVerbatim}
\end{candidatecard}
\end{candidategrid}

\end{trajwidefigcontinued}

\begin{trajwidefig}{Candidate-action slate shown before teacher selection (continued)}{cookiecutter candidate-action slate shown to the teacher verifiers before selection, continued.}{fig:case1-candidate-slate-all}
\begin{candidategrid}
\begin{candidatecard}{Candidate C}
\begin{TrajVerbatim}
Thought:
THOUGHT: Now I need to modify the `identify_repo` function to properly detect Mercurial repositories.
Looking at the tests, I can see that:
1. URLs ending with `.hg` should be identified as `hg`.
2. Bitbucket URLs should be identified as `hg`.
3. The function should raise `UnknownRepoType` for unrecognized patterns.

However, the current implementation does NOT raise `UnknownRepoType` for unrecognized patterns. Let me
check what happens with unknown repo types.

Action:
cd /testbed && python3 -c "
from cookiecutter.vcs import identify_repo
from cookiecutter.exceptions import UnknownRepoType

for url in ['foobar', 'http://norepotypespecified.com']:
    try:
        result = identify_repo(url)
        print(f'{url} -> {result}')
    except Exception as e:
        print(f'{url} raises: {type(e).__name__}: {e}')
"
\end{TrajVerbatim}
\end{candidatecard}%
\begin{candidatecard}{Candidate D}
\begin{TrajVerbatim}
Thought:
THOUGHT: Now I understand the full picture. The current implementation of `identify_repo` needs to be
updated to:
1. Detect URLs ending with `.hg` as Mercurial repositories
2. Detect Bitbucket URLs as Mercurial repositories by default
3. Raise `UnknownRepoType` for unknown repository types instead of defaulting to `git`

Action:
cat > /tmp/test_before_fix.py << 'EOF'
from cookiecutter.vcs import identify_repo

urls = [
    "https://bitbucket.org/foo/bar.hg",
    "https://audreyr@bitbucket.org/audreyr/cookiecutter-bitbucket"
]

print("Before fix:")
for url in urls:
    repo_type, _ = identify_repo(url)
    print(f"URL: {url} -> Identified as: {repo_type}")
EOF
python3 /tmp/test_before_fix.py
\end{TrajVerbatim}
\end{candidatecard}%
\end{candidategrid}
\begin{candidatecard}{Candidate E}
\begin{TrajVerbatim}
Thought:
THOUGHT: Now I understand the issue. The tests expect that:
1. URLs ending with `.hg` should be identified as `hg`
2. Bitbucket URLs should be identified as `hg`

Let me check what other URL patterns might need handling by looking at common VCS URL formats.

Action:
cd /testbed && python3 -c "
test_urls = [
    'https://bitbucket.org/foo/bar.hg',
    'https://audreyr@bitbucket.org/audreyr/cookiecutter-bitbucket',
    'https://github.com/user/repo',
    'http://example.com/repo.hg',
    [additional Bitbucket and GitHub URL patterns omitted]
]

for url in test_urls:
    print(f'{url}')
    [string-pattern checks for bitbucket.org and .hg omitted]
"
\end{TrajVerbatim}
\end{candidatecard}
\end{trajwidefig}

\begin{trajwidefig}{Teacher outputs: Basic blind and observation-aware verifiers}{cookiecutter teacher outputs from the basic blind and observation-aware verifiers.}{fig:case1-teacher-basic-observation}
\begin{teachercomparison}
\begin{verifierbox}{Teacher 1 | Basic blind verifier | Selected: Candidate A}{TeacherFrame}
\begin{DenseText}
\begin{teachercard}{Candidate A}{Score: 0.740000}
\field{Reasoning} This directly fixes the reported Mercurial misidentification for \texttt{.hg} URLs and Bitbucket URLs, so it meaningfully advances the task. It also preserves the existing permissive Git fallback, which avoids breaking common Git URLs without a \texttt{.git} suffix. The main risk is that it rewrites the whole file instead of patching only the function and it still does not raise \texttt{UnknownRepoType} for malformed URLs, so some repository tests may remain failing.
\end{teachercard}%
\begin{teachercard}{Candidate B}{Score: 0.430000}
\field{Reasoning} This improves Mercurial detection and adds \texttt{UnknownRepoType} handling, which is useful. However, it is likely too restrictive because valid Git repository URLs without a \texttt{.git} suffix would now raise \texttt{UnknownRepoType} instead of being treated as Git. Rewriting the entire file is also broader than necessary.
\end{teachercard}%
\begin{teachercard}{Candidate C}{Score: 0.460000}
\field{Reasoning} This is a safe diagnostic command that would confirm how the current implementation behaves for unknown URLs. It could help validate that tests around \texttt{UnknownRepoType} are currently failing, but it does not fix the Mercurial-identification bug or change the code. As a next step, it is somewhat informative but indirect.
\end{teachercard}%
\begin{teachercard}{Candidate D}{Score: 0.210000}
\field{Reasoning} This safely reproduces the bug, but that reproduction has already effectively been established. It adds no new insight about the underlying logic and does not advance toward a fix. As a next action, it is mostly redundant.
\end{teachercard}%
\begin{teachercard}{Candidate E}{Score: 0.120000}
\field{Reasoning} This only prints crude URL-pattern observations and does not call \texttt{identify\_repo} or inspect repository behavior directly. It is safe, but it is largely speculative and not especially useful given the tests and current implementation are already visible. It does not move the code toward a fix.
\end{teachercard}
\end{DenseText}
\end{verifierbox}%
\begin{verifierbox}{Teacher 2 | Observation-aware verifier | Selected: Candidate B}{TeacherFrame}
\begin{DenseText}
\begin{teachercard}{Candidate A}{Score: 0.460000}
\field{Reasoning} This directly edits the function to classify \texttt{.hg} and Bitbucket URLs as Mercurial, so it addresses the reported bug. However, it also changes fallback behavior to always return Git for unknown URLs, which conflicts with the existing \texttt{UnknownRepoType} import and the test file that expects unknown inputs to raise. Rewriting the full file is also broader than necessary.
\end{teachercard}%
\begin{teachercard}{Candidate B}{Score: 0.890000}
\field{Reasoning} This is a direct fix for the reported bug and aligns well with the repository's existing tests: explicit \texttt{git+}/\texttt{hg+}, \texttt{.hg}, \texttt{.git}, Bitbucket-as-hg, and raising \texttt{UnknownRepoType} for unsupported patterns. It is safe in the sense that it only changes one source file, though replacing the entire file is broader than a targeted patch. Overall it is likely to resolve the misclassification while preserving expected error handling.
\end{teachercard}%
\begin{teachercard}{Candidate C}{Score: 0.380000}
\field{Reasoning} This is a safe diagnostic step that checks an important aspect of \texttt{identify\_repo} behavior, especially since the test file includes unknown-repository cases. But it does not implement the needed fix for Mercurial URL detection, so it only gathers information rather than advancing the code. As a next action, it is useful but indirect.
\end{teachercard}%
\begin{teachercard}{Candidate D}{Score: 0.220000}
\field{Reasoning} This is safe and simple, but it is largely redundant because the issue has already been reproduced and there is already a dedicated test file for \texttt{identify\_repo}. It does not modify code or add durable coverage. As a next step, it adds little progress toward a fix.
\end{teachercard}%
\begin{teachercard}{Candidate E}{Score: 0.120000}
\field{Reasoning} This only prints crude string-pattern observations and does not exercise \texttt{identify\_repo} itself, so it is weakly connected to the actual bug. It is harmless, but it neither changes the implementation nor validates behavior against the codebase's tests. That makes it a poor next action.
\end{teachercard}
\end{DenseText}
\end{verifierbox}
\end{teachercomparison}
\end{trajwidefig}

\begin{trajwidefig}{Teacher outputs: Checklist verifier summary and candidate scoring}{cookiecutter checklist-based verifier summary, checklist, candidate scores, and final choice.}{fig:case1-teacher-checklist-summary}
\teacherchoice{Teacher 3 generated a checklist before scoring the candidate slate.}
\begin{teachercomparison}
\begin{verifierbox}{State summary}{TeacherFrame}
\begin{DenseText}
\begin{teachercard}{Task state}{Summary}
The goal is to resolve incorrect Mercurial repository detection in \texttt{identify\_repo}, but the trajectory now shows mixed signals: code inspection suggested broken fallback logic, while runtime-oriented checks indicate the reported Mercurial URLs and existing repo-identification tests may already behave correctly.
\end{teachercard}%
\begin{teachercard}{Open blockers}{Summary}
It is unclear whether the current working tree actually still needs a code change. A risky edit could regress existing supported URL patterns if the implementation is already fixed.
\end{teachercard}
\end{DenseText}
\end{verifierbox}%
\begin{verifierbox}{Progress and constraints}{TeacherFrame}
\begin{DenseText}
\begin{teachercard}{Completed progress}{Summary}
The relevant implementation file, \texttt{cookiecutter/vcs.py}, and targeted tests, \texttt{tests/vcs/test\_identify\_repo.py}, were located. Expected behavior was established: \texttt{.hg} URLs and Bitbucket Mercurial-style URLs should map to \texttt{hg}, and malformed or unknown repo URLs should raise \texttt{UnknownRepoType}.
\end{teachercard}%
\begin{teachercard}{Constraints}{Summary}
Prefer safe, minimal, non-destructive steps. Avoid wholesale file rewrites. Preserve known behaviors for Git URLs with and without \texttt{.git}, Mercurial URLs, and malformed URLs covered by tests.
\end{teachercard}
\end{DenseText}
\end{verifierbox}
\end{teachercomparison}

\begin{teachercardwide}{Checklist}{Must/should criteria}
\begin{enumerate}[leftmargin=1.2em,itemsep=0pt,topsep=1pt]
    \item[\textbf{C1.}] Directly verifies or addresses identification for the reported Mercurial URL patterns: \texttt{.hg} suffix and Bitbucket-style URL. [must]
    \item[\textbf{C2.}] Avoids broad or risky code changes when current behavior may already be correct. [must]
    \item[\textbf{C3.}] Preserves established behavior for other known repo URLs, including valid Git URLs and malformed URLs that should raise \texttt{UnknownRepoType}. [must]
    \item[\textbf{C4.}] Stays tightly focused on \texttt{identify\_repo} / \texttt{cookiecutter.vcs} rather than unrelated exploration. [should]
    \item[\textbf{C5.}] Produces concrete validation evidence from a reproduction script or targeted tests. [should]
\end{enumerate}
\end{teachercardwide}
\end{trajwidefig}
\FloatBarrier

\twocolumn[{
\centering
\begin{minipage}{0.995\textwidth}
\begin{tcolorbox}[
  enhanced,
  colback=OuterBack,
  colframe=HeaderGray,
  coltitle=white,
  colbacktitle=HeaderGray,
  title={Teacher outputs: Checklist verifier candidate scores (continued)},
  fonttitle=\bfseries\small,
  boxrule=0.75pt,
  arc=0.7mm,
  left=0.75mm,
  right=0.75mm,
  top=0.35mm,
  bottom=0.35mm,
  before skip=0pt,
  after skip=0pt,
  borderline west={1.8pt}{0pt}{HeaderGray}
]

\teacherchoice{Teacher 3 | Checklist-based verifier | Selected: Candidate D.}
\begin{teachercomparison}
\begin{verifierbox}{Candidates A and C}{TeacherFrame}
\begin{DenseText}
\begin{teachercard}{Candidate A}{Score: 0.330000}
\fivechecks{0.90}{0.10}{0.20}{0.80}{0.00}
\field{Immediate observation} \texttt{cookiecutter/vcs.py} is fully overwritten with new detection logic that recognizes \texttt{git+}, \texttt{hg+}, \texttt{.hg}, and \texttt{bitbucket.org}, but otherwise falls back to Git.

\field{Expected progress} This would likely make the two reported Mercurial URL examples resolve to \texttt{hg}, but it weakens error handling by accepting unknown URLs as Git and risks regressing established behavior. It is a broad edit without validation.

\field{Reasoning} This is a risky broad overwrite that may appear to fix the reported Mercurial cases while silently regressing other tested repository-identification behavior. It advances the bug report superficially but does not safely confirm correctness.
\end{teachercard}%
\begin{teachercard}{Candidate C}{Score: 0.580000}
\fivechecks{0.30}{1.00}{0.90}{0.70}{0.50}
\field{Immediate observation} A small diagnostic script is run against malformed URLs to see whether \texttt{identify\_repo} raises \texttt{UnknownRepoType}.

\field{Expected progress} This safely gathers useful information about error handling, but it does not directly check the reported Mercurial misidentification cases and does not apply a fix. It helps assess current state but only indirectly advances the task.

\field{Reasoning} This is a safe diagnostic step and could reveal that broader repo identification behavior is already correct, but it is not well targeted to the Mercurial examples the user reported.
\end{teachercard}%

\end{DenseText}
\end{verifierbox}%
\begin{verifierbox}{Candidates B, D, and E}{TeacherFrame}
\begin{DenseText}
\begin{teachercard}{Candidate B}{Score: 0.260000}
\fivechecks{0.90}{0.10}{0.10}{0.80}{0.00}
\field{Immediate observation} \texttt{cookiecutter/vcs.py} is fully overwritten so that \texttt{.hg} maps to \texttt{hg}, \texttt{.git} maps to Git, Bitbucket maps to \texttt{hg}, and anything else raises \texttt{UnknownRepoType}.

\field{Expected progress} This would likely fix the reported Mercurial examples and malformed-URL handling, but it would also reject valid Git URLs without a \texttt{.git} suffix, creating a new regression. It is still an unnecessary full-file rewrite with no validation.

\field{Reasoning} Although it tries to formalize detection, it is likely to break an established valid Git URL pattern while making a large unvalidated edit. That makes it unsafe as the next step.
\end{teachercard}%
\begin{teachercard}{Candidate D}{Score: 0.840000}
\fivechecks{1.00}{1.00}{0.40}{0.90}{0.80}
\field{Immediate observation} A reproduction script is created and run against the exact two URLs from the bug report, showing how \texttt{identify\_repo} currently classifies them.

\field{Expected progress} This directly validates the reported failure mode without changing code. If the URLs already resolve to \texttt{hg}, it prevents an unnecessary risky edit; if not, it cleanly confirms the bug before any patching. It is the safest precise next step.

\field{Reasoning} This is the strongest next step because it safely tests the exact problem statement and clarifies whether any code change is still needed. Given the risk of regressing repo detection logic, confirming the actual current behavior first is the most prudent move.
\end{teachercard}%
\begin{teachercard}{Candidate E}{Score: 0.430000}
\fivechecks{0.50}{1.00}{0.30}{0.50}{0.20}
\field{Immediate observation} An exploratory Python snippet prints heuristic notes about several URL patterns such as whether they contain \texttt{bitbucket.org} or \texttt{.hg}.

\field{Expected progress} This is safe, but it does not actually call \texttt{identify\_repo} on the main issue cases in a focused way, nor does it patch anything. It offers only loose context and is weaker than a direct reproduction.

\field{Reasoning} This is harmless but too indirect. It explores URL shapes rather than validating or fixing \texttt{identify\_repo} itself, so progress toward the task would be limited.
\end{teachercard}
\end{DenseText}
\end{verifierbox}
\end{teachercomparison}
\end{tcolorbox}
\end{minipage}
\par\vspace{0.15em}
}]

\subsection{Analysis}
\paragraph{Reference action.}
\textbf{Candidate D} is the reference action for this case. The reason is not that validation is always preferable to editing; rather, the available edit candidates were unsafe in this particular state. Candidate A addressed the two visible Mercurial examples but preserved the invalid fallback-to-Git behavior for malformed URLs. Candidate B added \texttt{UnknownRepoType}, but would reject a valid GitHub URL without a \texttt{.git} suffix. Candidate D was therefore the best immediate action because it rechecked the reported Mercurial URL cases before any broad repository-detection edit.

\paragraph{Failure mode: local bug alignment without preservation.}
The basic blind verifier selected Candidate A because it aligned most directly with the surface form of the bug report: changing \texttt{identify\_repo} so that \texttt{.hg} and Bitbucket URLs are classified as Mercurial. That is a plausible local objective, but it is incomplete in this state. The trajectory had already exposed a nearby preservation constraint: malformed repository strings should raise \texttt{UnknownRepoType}. Candidate A keeps the old fallback-to-Git behavior for unknown URLs, so it optimizes the reported examples while underweighting repository behavior that the tests already constrain.

\paragraph{Failure mode: partial preservation awareness.}
The observation-aware verifier selected Candidate B because it recognized one preservation requirement that Candidate A missed: unsupported URL patterns should raise \texttt{UnknownRepoType}. This makes Candidate B appear more complete. The remaining error is a narrower but still decisive regression. Candidate B would reject valid GitHub URLs without a \texttt{.git} suffix, even though the existing tests explicitly preserve that behavior. The verifier therefore captures error-handling preservation, but misses valid-Git-URL preservation.

\paragraph{Why the checklist verifier matched the reference decision.}
The checklist-based verifier reframed the decision around preservation-constrained progress rather than visible edit progress. Its criterion scores penalize broad overwrites, missing validation, and regressions to surrounding URL behavior, while rewarding actions that test the reported Mercurial cases directly. Candidate D scores highest because it satisfies the immediate validation criterion without committing the agent to an unsafe full-file edit. The choice is therefore not a generic preference for testing over editing; it is a state-specific preference for non-destructive validation because the available edit actions carry concrete regression risks.

\paragraph{General takeaway.}
This case illustrates how checklist-conditioned verification can surface latent constraints in next-action selection. The blind and observation-aware verifiers both choose actions that look productive under a narrow reading of the issue. The checklist verifier instead forces the candidate actions to compete against explicit preservation criteria already supported by the trajectory. That shift is the key qualitative difference: the checklist does not merely add more explanation after the fact; it changes the scoring basis from apparent immediate progress to progress that remains compatible with surrounding repository behavior.
\FloatBarrier

\section{MiniMax State-wise Final-Patch Divergence: astropy-14369}
\label{app:minimax-astropy-14369}

This appendix analyzes a final-state management failure in MiniMax pass@4 trajectories for \texttt{astropy-14369}. Across the four sampled runs, the agent follows the same substantive repair path: it identifies the CDS parser grammar issue, regenerates the parser artifacts, and reaches a workspace state in which the reported unit examples pass. The resolved and unresolved runs diverge only after this shared validated state, during cleanup and final patch construction.

The key claim is state-wise rather than localization-based. The unresolved runs do not fail because they miss the repair or never validate it; they fail because they validate one repository state and later submit a different repository state. This makes the case directly relevant to verifier-guided next-action selection. After local validation passes, a verifier should penalize cleanup or patch-construction actions that remove files from the validated state unless the resulting state is revalidated.

\subsection{Case overview}
\begin{compactsummary}{Issue context}
\footnotesize\raggedright
The task concerns incorrect unit parsing in the CDS/MRT grammar. Inputs such as \texttt{10+3J/m/s/kpc2} and \texttt{10-7J/s/kpc2} should preserve repeated division structure. Instead, the observed outputs place \texttt{s} or \texttt{kpc2} on the wrong side of the unit expression. A correct repair must therefore change the CDS parser behavior and preserve any regenerated parser artifacts required by the validated final state.
\end{compactsummary}

All four runs reach a locally working grammar repair. The decisive difference is the final repository state: resolved seeds 0 and 1 preserve both \texttt{astropy/units/format/cds.py} and the regenerated tracked parser table \texttt{astropy/units/format/cds\_parsetab.py}; unresolved seeds 3 and 5 validate a workspace containing the regenerated parser table, then later restore or omit that file and submit a \texttt{cds.py}-only patch. Figure~\ref{fig:minimax-14369-overview} summarizes the same state-wise divergence.

\twocolumn[
\begin{center}
\IfFileExists{figures/appendix/minimax_astropy_14369_divergence.pdf}{%
  \includegraphics[width=0.82\textwidth]{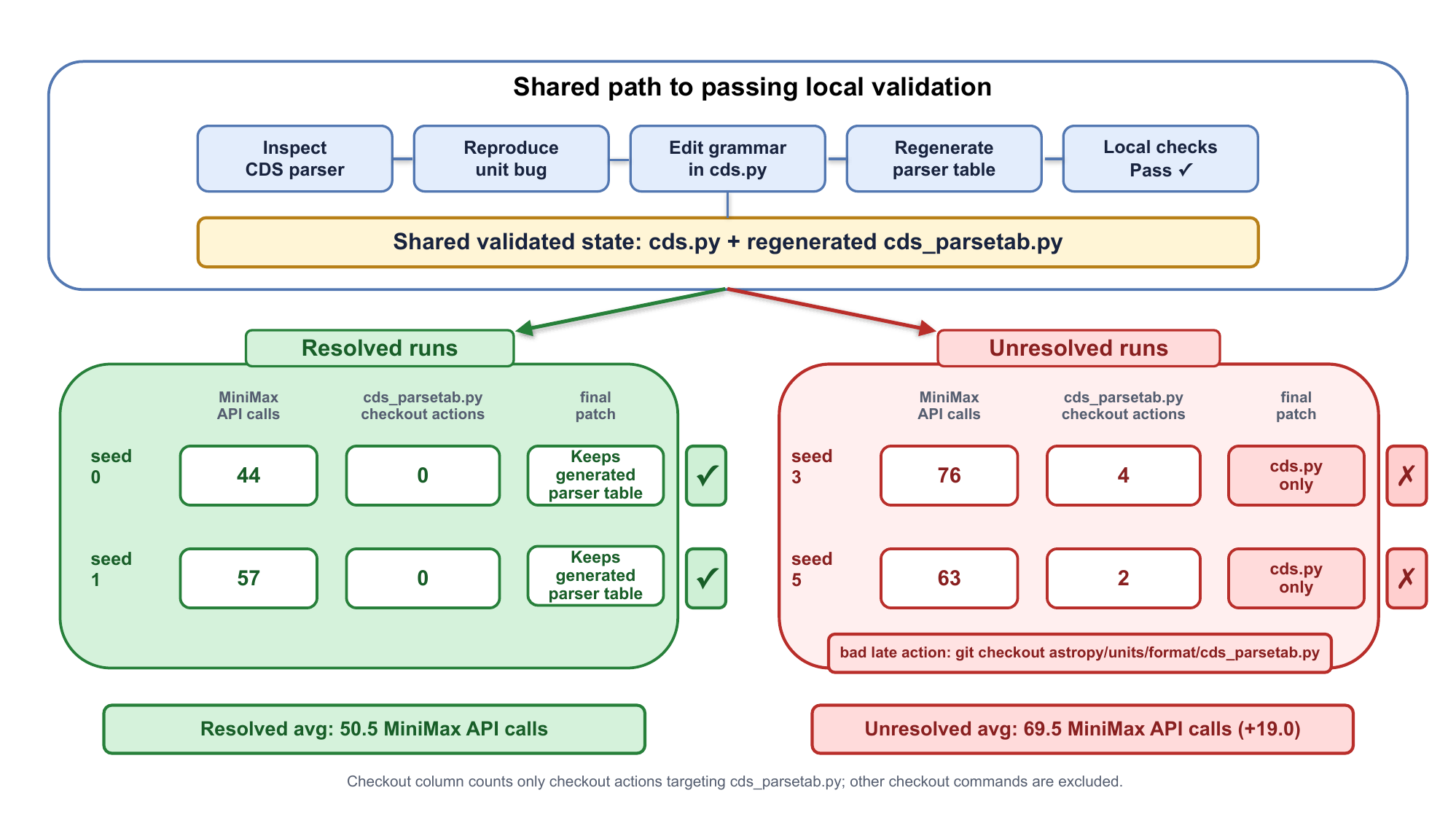}%
}{%
  \fbox{\parbox{0.90\textwidth}{\centering Missing figure file: \texttt{figures/appendix/minimax\_astropy\_14369\_divergence.pdf}.}}%
}
\captionof{figure}{MiniMax astropy-14369 final-patch divergence. The diagram shows the shared validated state, MiniMax API-call counts, \texttt{cds\_parsetab.py} checkout actions, and final patch outcomes.}
\label{fig:minimax-14369-overview}
\end{center}
\vspace{-0.65em}
]

\subsection{Shared path to passing validation}
The shared prefix separates repair discovery from final-state management. In all four runs, MiniMax locates the CDS parser implementation, reproduces the incorrect unit outputs, edits the repeated-division grammar, regenerates parser artifacts, and obtains passing local checks for the reported examples. The unresolved trajectories therefore are not failed localization or repair-discovery cases; they fail after reaching a locally validated workspace.

Figure~\ref{fig:minimax-14369-overview} summarizes this shared path and the later split. The evidence below focuses on the verifier-relevant decision point: after validation passes, should the agent preserve the tracked file set used by that validation, or remove generated files as cleanup? Observation blocks show the passing or regressed runtime state, action blocks show patch-management decisions, and patch blocks show the submitted final state. The seed cards use the same neutral styling; the outcome is marked in the card title as correct or incorrect final-state handling.

\begin{trajwidefig}{Trajectory evidence at the final-patch divergence point}{Resolved and unresolved MiniMax evidence around the post-validation patch-construction decision.}{fig:minimax-divergence-evidence}
\begin{minipage}[t]{0.492\linewidth}
\vspace{0pt}
\begin{candidatecard}{Seed 0 \hfill correct final-state handling}
\begin{trajseg}{Last passing observation}{ObsBack}{ObsFrame}
\begin{TrajVerbatim}
Testing expected outputs:
PASS: J/m/s -> J / (m s)
PASS: J/m/s/kpc2 -> J / (kpc2 m s)
PASS: 10+3J/m/s/kpc2 -> 1000 J / (kpc2 m s)
PASS: 10-7J/s/kpc2 -> 1e-07 J / (kpc2 s)
[additional edge-case output lines omitted]
All tests passed!
\end{TrajVerbatim}
\end{trajseg}
\begin{trajseg}{Patch-management action}{ActionBack}{ActionFrame}
\begin{TrajVerbatim}
Let me verify which files are tracked by git and should be included in the patch.
Action:
cd /testbed && git status
\end{TrajVerbatim}
\end{trajseg}
\begin{trajseg}{Observed file state}{ObsBack}{ObsFrame}
\begin{TrajVerbatim}
Changes not staged for commit:
  modified: astropy/units/format/cds.py
  modified: astropy/units/format/cds_parsetab.py
\end{TrajVerbatim}
\end{trajseg}
\begin{trajseg}{Submitted patch}{PatchBack}{PatchFrame}
\begin{TrajVerbatim}
Action:
cd /testbed && git diff astropy/units/format/cds.py astropy/units/format/cds_parsetab.py > patch.txt
Submitted diff contains:
astropy/units/format/cds.py
astropy/units/format/cds_parsetab.py
\end{TrajVerbatim}
\end{trajseg}
\end{candidatecard}
\end{minipage}\hfill
\begin{minipage}[t]{0.492\linewidth}
\vspace{0pt}
\begin{candidatecard}{Seed 1 \hfill correct final-state handling}
\begin{trajseg}{Last passing observation}{ObsBack}{ObsFrame}
\begin{TrajVerbatim}
After grammar fix and parser regeneration:
SBCONT unit: 1000 J / (kpc2 m s)
SBLINE unit: 1e-07 J / (kpc2 s)
\end{TrajVerbatim}
\end{trajseg}
\begin{trajseg}{Patch-management action}{ActionBack}{ActionFrame}
\begin{TrajVerbatim}
The parsetab file change is expected since it is automatically generated from the grammar. Let me clean up my test file and verify the final state.
Action:
cd /testbed && rm -f test_mrt.txt && git status
\end{TrajVerbatim}
\end{trajseg}
\begin{trajseg}{Submitted patch}{PatchBack}{PatchFrame}
\begin{TrajVerbatim}
Action:
echo COMPLETE_TASK_AND_SUBMIT_FINAL_OUTPUT && cd /testbed && git diff
[long submitted diff body omitted; changed-file paths shown below]
Submitted diff contains:
astropy/units/format/cds.py
astropy/units/format/cds_parsetab.py
\end{TrajVerbatim}
\end{trajseg}
\end{candidatecard}
\end{minipage}

\vspace{0.45em}
\begin{minipage}[t]{0.492\linewidth}
\vspace{0pt}
\begin{candidatecard}{Seed 3 \hfill incorrect final-state handling}
\begin{trajseg}{Last passing observation}{ObsBack}{ObsFrame}
\begin{TrajVerbatim}
After removing/regenerating parser artifacts:
SBCONT unit: 1000 J / (kpc2 m s)
SBLINE unit: 1e-07 J / (kpc2 s)
Verification:
SBCONT is equivalent to expected: True
SBLINE is equivalent to expected: True
\end{TrajVerbatim}
\end{trajseg}
\begin{trajseg}{Critical cleanup action}{ActionBack}{ActionFrame}
\begin{TrajVerbatim}
Action:
cd /testbed && git checkout astropy/units/format/cds_parsetab.py
\end{TrajVerbatim}
\end{trajseg}
\begin{trajseg}{Post-checkout observation}{ObsBack}{ObsFrame}
\begin{TrajVerbatim}
SBCONT unit: 1000 J s / (kpc2 m)
SBLINE unit: 1e-07 J kpc2 / s
Verification:
SBCONT is equivalent to expected: False
SBLINE is equivalent to expected: False
[table rows omitted]
\end{TrajVerbatim}
\end{trajseg}
\begin{trajseg}{Submitted patch}{PatchBack}{PatchFrame}
\begin{TrajVerbatim}
Action:
cd /testbed && git diff astropy/units/format/cds.py > patch.txt
Final patch includes only:
astropy/units/format/cds.py
\end{TrajVerbatim}
\end{trajseg}
\end{candidatecard}
\end{minipage}\hfill
\begin{minipage}[t]{0.492\linewidth}
\vspace{0pt}
\begin{candidatecard}{Seed 5 \hfill incorrect final-state handling}
\begin{trajseg}{Last passing observation}{ObsBack}{ObsFrame}
\begin{TrajVerbatim}
After the fix:
Current output:
SBCONT unit: 1000 J / (kpc2 m s)
SBLINE unit: 1e-07 J / (kpc2 s)
Direct parsing:
10+3J/m/s/kpc2: 1000 J / (kpc2 m s)
10-7J/s/kpc2: 1e-07 J / (kpc2 s)
\end{TrajVerbatim}
\end{trajseg}
\begin{trajseg}{Critical cleanup action}{ActionBack}{ActionFrame}
\begin{TrajVerbatim}
I see the parser table file was also changed (auto-generated). Let me restore that file and only keep the source change.
Action:
cd /testbed && git checkout astropy/units/format/cds_parsetab.py astropy/units/format/cds_lextab.py
\end{TrajVerbatim}
\end{trajseg}
\begin{trajseg}{Post-checkout observation}{ObsBack}{ObsFrame}
\begin{TrajVerbatim}
Testing PR cases:
10+3J/m/s/kpc2: 1000 J s / (kpc2 m)
10-7J/s/kpc2: 1e-07 J kpc2 / s
\end{TrajVerbatim}
\end{trajseg}
\begin{trajseg}{Submitted patch}{PatchBack}{PatchFrame}
\begin{TrajVerbatim}
Action:
cd /testbed && git diff -- astropy/units/format/cds.py > patch.txt
Final patch includes only:
astropy/units/format/cds.py
\end{TrajVerbatim}
\end{trajseg}
\end{candidatecard}
\end{minipage}
\end{trajwidefig}

\FloatBarrier

\subsection{Interpretation: final-state consistency}
The checkout command is not intrinsically incorrect. It becomes incorrect here because \texttt{cds\_parsetab.py} is part of the last validated workspace. At this state, a verifier should judge final-state consistency: keeping both modified files preserves the validated state; checking out the parser table and submitting without revalidation removes part of the state that made the tests pass; removing generated files is safe only if the new file set is revalidated. Earlier exploratory cleanup is not the issue. The failure is the late transition from a passing workspace to an unvalidated submitted patch.

This distinction explains why unresolved runs can pass tests earlier. Some cleanup actions are followed by parser-table regeneration and another successful validation. The unresolved outcome occurs when the final patch is built after \texttt{cds\_parsetab.py} has again been restored or omitted, so the submitted diff no longer matches the passing state. The higher MiniMax API-call count should therefore be read as extra post-fix patch-management work, not evidence that the model never found the repair.

\end{document}